\documentclass{article}

\usepackage{iclr2025_conference}
\usepackage{times}

\iclrfinalcopy 

\usepackage{tablefootnote}

\usepackage[utf8]{inputenc} %
\usepackage{makecell}
\usepackage{caption}
\usepackage{hyperref}       %
\usepackage{url}            %
\usepackage{booktabs}       %
\usepackage{amsfonts}       %
\usepackage{nicefrac}       %
\usepackage{enumitem}
\usepackage{microtype}      %
\usepackage{xcolor}         %
\usepackage{multirow}
\usepackage{graphicx}
\usepackage{tikz}
\usepackage{xr}
\usepackage{amsmath}
\usepackage{amssymb}
\usepackage{xspace}
\usepackage{colortbl}
\usepackage{subcaption}
\usepackage[accsupp]{axessibility}

\let\cite\citep

\title{%
A Preliminary Study of \texttt{o1} in Medicine:\\ Are We Closer to an AI Doctor \includegraphics[width=.6cm]{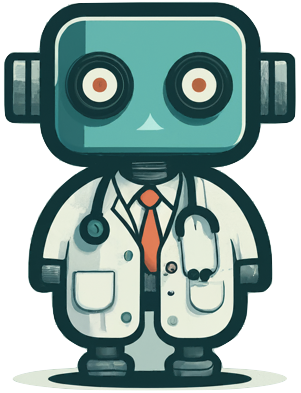}?}

\author{%
\bf Yunfei Xie$^{1\ast}$ \, Juncheng Wu$^{1\ast}$ \, Haoqin Tu$^{1\ast}$ \, Siwei Yang$^{1\ast}$ \vspace{.15em} \\  \bf  Bingchen Zhao$^2$ \,
Yongshuo Zong$^2$ \,  Qiao Jin$^3$ \, Cihang Xie$^1$ \, Yuyin Zhou$^1$ 
\vspace{.3em} \\
$^{\star}$equal technical contribution\vspace{.5em} \\
$^1$UC Santa Cruz \quad \quad
$^2$University of Edinburgh \quad \quad
$^3$National Institutes of Health
}

\usepackage{enumitem}
\usepackage[most]{tcolorbox}

 \makeatletter\renewcommand\paragraph{\@startsection{paragraph}{4}{\z@}
	{.2em \@plus1ex \@minus.2ex}{-.5em}{\normalfont\normalsize\bfseries}}\makeatother
\makeatletter
\DeclareRobustCommand\onedot{\futurelet\@let@token\@onedot}
\def\@onedot{\ifx\@let@token.\else.\null\fi\xspace}
\definecolor{ForestGreen}{RGB}{34,139,34}
\def\eg{\emph{e.g}\onedot} 
\def\ie{\emph{i.e}\onedot}

\def\vs{\emph{vs}\onedot}

\makeatother

\definecolor{baselinecolor}{gray}{.9}
\newcommand{\bl}[1]{\cellcolor{baselinecolor}{#1}}

\makeatletter
\let\@oldmaketitle\@maketitle%
\renewcommand{\@maketitle}{\@oldmaketitle%
     \begin{figure*}[htbp]
  \centering
  \begin{minipage}[b]{0.55\textwidth}
    \centering
    \includegraphics[width=\textwidth]{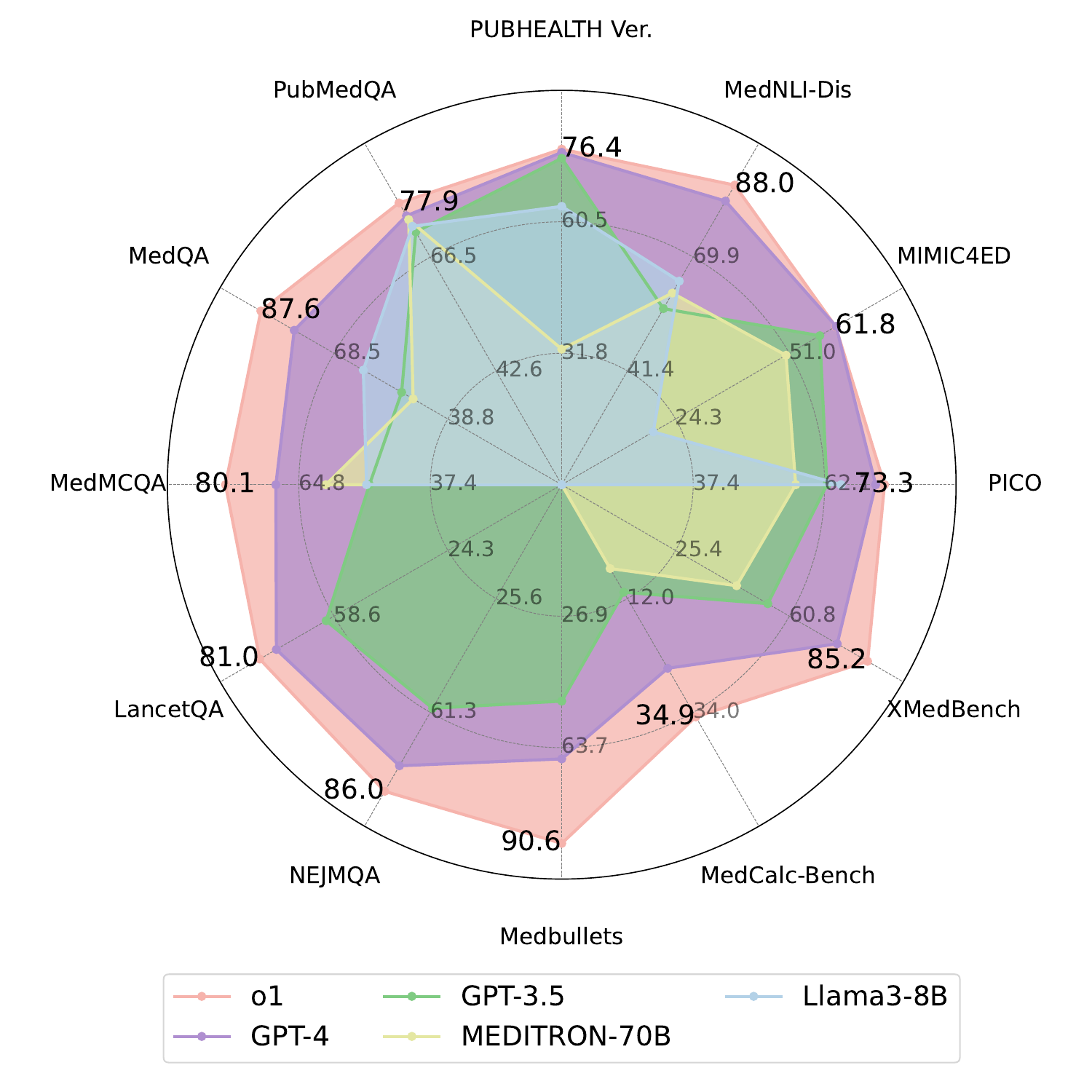}
    \caption{Overall results of \texttt{o1} and other 4 strong LLMs. We show performance on 12 medical datasets spanning diverse domains. \texttt{o1} demonstrates a clear performance advantage over close- and open-source models.}
    \label{fig:radar}
  \end{minipage}
  \hfill
  \begin{minipage}[b]{0.4\textwidth}
    \centering

    \includegraphics[width=\textwidth]{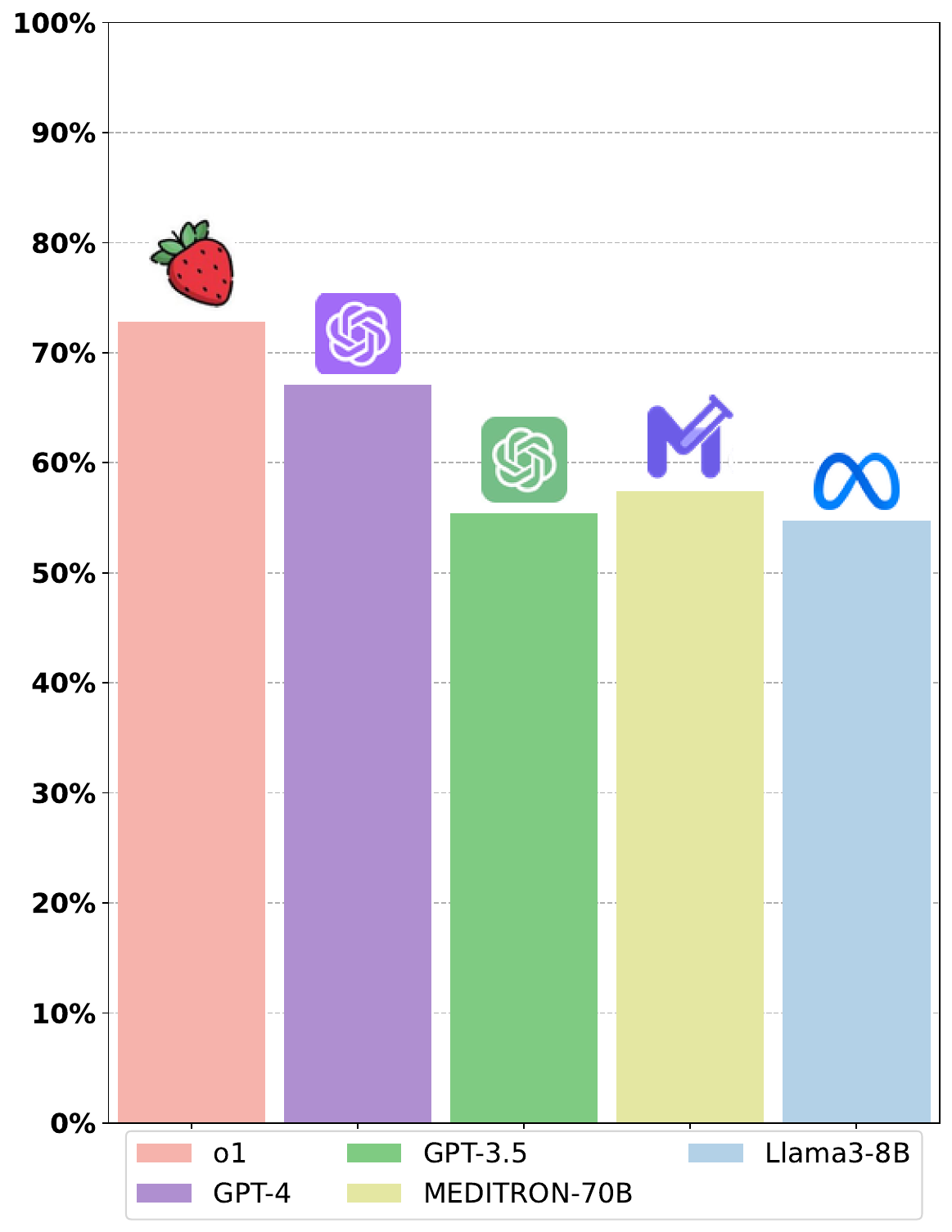}
    \caption{Average accuracy of \texttt{o1} and other 4 strong LLMs. \texttt{o1} achieves the highest average accuracy of 74.3\% across 19 medical datasets.}
    \label{fig:bar}
  \end{minipage}

  \label{fig:radar_and_bar}
\end{figure*}

    \bigskip}%
\makeatother

\begin{document}

\maketitle

\begin{abstract}
  Large language models (LLMs) have exhibited remarkable capabilities across various domains and tasks, pushing the boundaries of our knowledge in learning and cognition. The latest model, OpenAI's \texttt{o1}, stands out as the first LLM with an internalized chain-of-thought technique using reinforcement learning strategies. While it has demonstrated surprisingly strong capabilities on various general language tasks, its performance in specialized fields such as medicine remains unknown. To this end, this report provides a preliminary exploration of \texttt{o1} on different medical scenarios, comprehensively examining 3 key aspects: \emph{understanding}, \emph{reasoning}, and \emph{multilinguality}. Specifically, our evaluation encompasses 6 tasks using data from 37 medical datasets, including two newly constructed and more challenging question-answering (QA) tasks based on professional medical quizzes from the New England Journal of Medicine and The Lancet. These datasets offer greater clinical relevance compared to standard medical QA benchmarks such as MedQA, translating more effectively into real-world clinical utility. Our analysis of \texttt{o1} suggests that the enhanced reasoning ability of LLMs may (significantly) benefit their capability to understand various medical instructions and reason through complex clinical scenarios. Notably, \texttt{o1} surpasses the previous GPT-4 in accuracy by an average of 6.2\% and 6.6\% across 19 datasets and two newly created complex QA scenarios. But meanwhile, we also identify several weaknesses in both the model capability and the existing evaluation protocols, including hallucination, inconsistent multilingual ability, and discrepant metrics for evaluation. We release our raw data and model outputs at \url{https://ucsc-vlaa.github.io/o1_medicine/} for future research.

\end{abstract}

\section{Introduction}

Intelligence, a complex and elusive concept, has puzzled psychologists, philosophers, and computer scientists for years~\cite{bubeck2023sparks}. 
While there is no single agreed-upon definition of intelligence, it is widely accepted that it spans a broad range of cognitive skills, rather than being confined to a specific task~\cite{mccarthy2006proposal}. Creating artificial systems with such general intelligence has been a long-standing and ambitious goal of AI research. 
The most exciting progresses in AI are achieved by language models in these years, from the initial start of ChatGPT to its evolution and other open-source projects~\cite{touvron2023llama,touvron2023llama2,jiang2023mistral,bai2023qwen,rwkv6}. 

Early LLM pioneers set out goals to understand and interact with human by exploring generalizable reasoning mechanisms and building knowledge bases with vast amounts of commonsense information. With parameters and data volume in place, the question of how to effectively prompt the model from the user end and train it from the developer end has become a trending topic of exploration~\cite{wei2022chain,ouyang2022training}. On the user side, varying prompting techniques can significantly impact model performance. Chain-of-thought (CoT) prompting~\cite{wei2022chain,dong2022survey,saunders2022self}, one of the most popular strategies, leverages the model’s internal reasoning patterns to enhance its ability to solve complex tasks. OpenAI capitalized on this by embedding the CoT process into model training, integrating reinforcement learning, and finally introduced the \texttt{o1} model~\cite{openai-o1}. 
While the \texttt{o1} model demonstrates strong performance in general domains, its effectiveness in specialized fields like medicine---where domain-specific training may be lacking---remains uncertain. Moreover, current benchmarks for LLMs in the medical domain often evaluate models only on a limited set of factors, often focusing on isolated aspects such as knowledge and reasoning~\cite{nori2023can,lievin2024can}, safety~\cite{han2024towards}, or multilinguality~\cite{wang2024apollo}.
These factors make a comprehensive assessment of LLMs' capabilities---especially for advanced models like \texttt{o1}---in medical challenging tasks (Figure~\ref{fig:radar}).

This paper aims to provide an initiative to close this gap, focusing on \texttt{o1}.
We identify three fundamental aspects of LLMs in medicine: \emph{understanding}, \emph{reasoning}, and \emph{multilinguality}. 
To evaluate these capabilities, we assembled 35 existing medical datasets and developed two novel, challenging QA datasets that include instructions and expected outputs, ensuring comprehensive assessment. With evaluation on this extensive suite, our key findings include: 
\begin{itemize} [leftmargin=12pt]
    \item \texttt{o1} demonstrates improved transfer of clinical understanding and reasoning abilities, validating its competence in real-world diagnostic scenarios compared with both close- and open-source models as presented in Figure~\ref{fig:radar} and Figure~\ref{fig:bar};
    \item No single model excels across all tasks on our medical leaderboard, though \texttt{o1} comes close to dominating most evaluations;
    \item  \texttt{o1} still suffers from the long-standing issue of hallucination and complex multilingual medical cases;
    \item Inconsistencies in metrics for medical NLP can significantly affect models' standings, which calls for a re-evaluation of reliable metrics for future LLMs;
    \item CoT prompting can further enhance \texttt{o1} in medicine, despite its training having already integrated CoT data.
\end{itemize}
In addition to these findings, we also elevate the discussion section as an initial attempt to address the issues identified during our benchmarking in Section~\ref{sec:discussion}. 
Particularly, we highlight the potential negative effects of \texttt{o1}, emphasize the urgent need for consistent and unified evaluation metrics for future LLMs, and advocate for improved instruction templates that can be applied to models with embedded prompting strategies.

\section{Related works}
\paragraph{Large Language Models with Enhanced Reasoning Ability.}
Large Language models (LLMs) based on next token prediction pre-training~\cite{touvron2023llama,touvron2023llama2,achiam2023gpt} have demonstrated promising capabilities on various language undersanding tasks. 
Instruction fine-tuning further improved the abilites of these LLMs for following user instructions.
However, recent studies suggest that LLMs struggle with complex tasks involving logical reasoning.
To address this issue, some researches propose to instruct LLMs to mimic human thinking processes by producing a chain-of-thought (CoT)~\cite{feng2024towards,wei2022chain} before generating a final answer.
Reinforcement learning from human feedback~\cite{ouyang2022training} has also been employed to enhance reasoning while make sure the models align with human values~\cite{tu2023sight,tu2023how}.
Recently, OpenAI introduced \texttt{o1}, which was trained on a vast amount of CoT data, further enhancing the capability of LLMs in solving scientific problems.
In this paper, we aim to investigate whether enhanced abilities of \texttt{o1} effectively transfer to the clinical medical domain.

\paragraph{Medical Large Language Models.}
 Benefiting from the generalization capabilities of LLMs, general-purpose models such as GPT-4 have demonstrated impressive performance on challenging medical problems~\cite{nori2023capabilities,wu2024towards}.
Some researchers have attempted to further equip LLMs with biomedical knowledge by fine-tuning them using domain-specific corpora~\cite{chen2023meditron,wang2023huatuo,wu2024pmc,li2023chatdoctor}.
However, for clinical applications, LLMs are not only required to understand medical domain-specific knowledge but also to produce reliable responses by performing logical reasoning.
In this paper, we aim to explore the potential of \texttt{o1} as a clinical viable model.
Our experimental findings reveal that with enhanced \emph{understanding}, \emph{reasoning}, and \emph{multilinguality} medical capabilities, \texttt{o1} makes a step closer to reliable clinical AI-system.

\section{Evaluation Pipeline}
\subsection{Overall Taxonomy of Evaluations}
\begin{figure*}[t]
  \centering
  \includegraphics[page=1, width=0.9\textwidth]{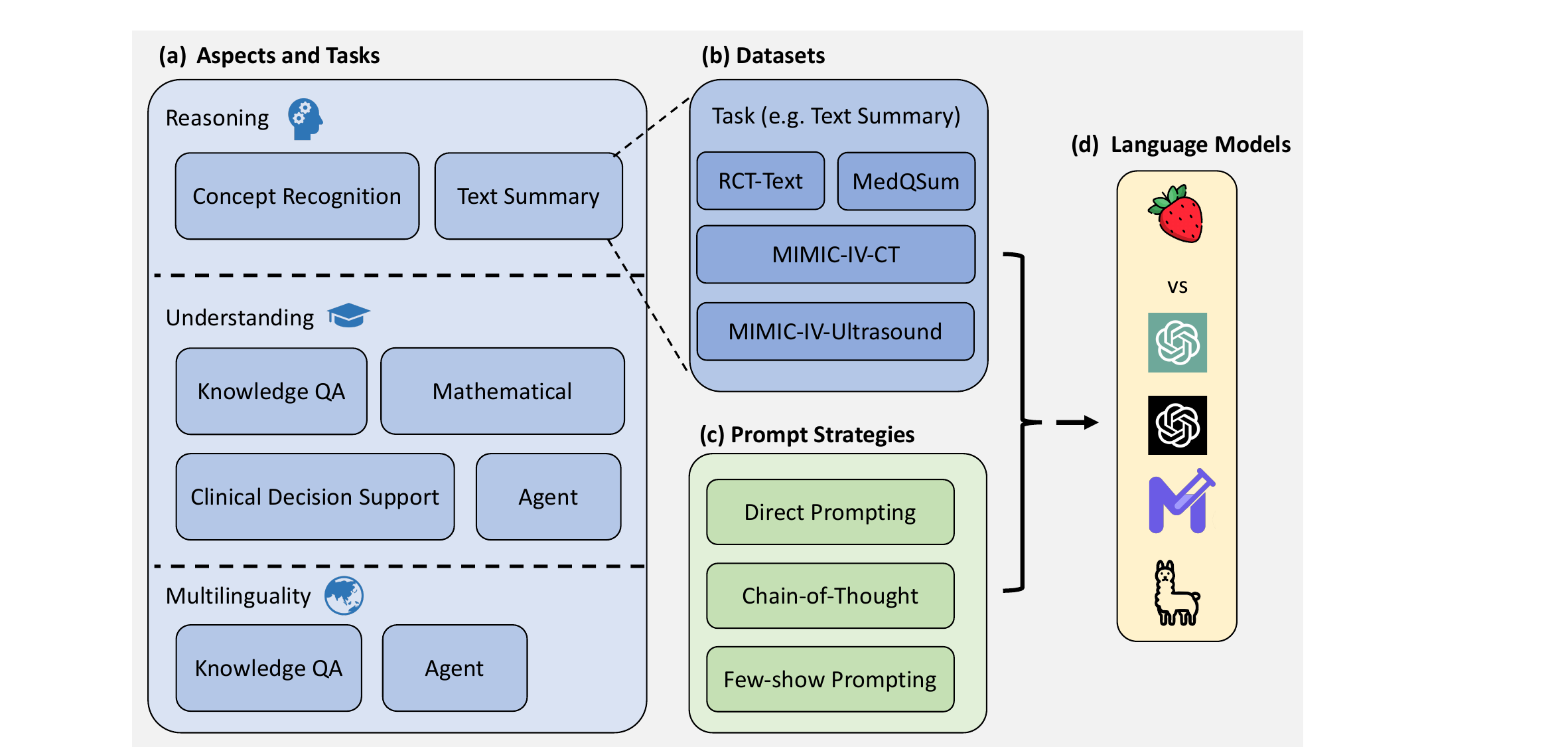}
  \caption{Our evaluation pipeline has different (a) \emph{aspects} with various (b) \emph{prompting strategies}  using the latest (c) \emph{language models}. We leverage a comprehensive set of (d) \emph{evaluations} to present a holistic view of model progress in the medical domain.}
  \label{fig:pipeline}
\end{figure*}

\begin{table}[htbp]
  \centering
  \setlength{\tabcolsep}{1.8pt}
  \caption{Six datasets across three fundamental aspects employed in our evaluation suite. Asterisks (*) denotes the newly constructed datasets from public sources.  %
  }\label{table:dataset_scenarios}
  \resizebox*{!}{.8\textheight}{
    \begin{tabular}{rcccp{18em}c}
          &       &       &       & \multicolumn{1}{r}{} &  \\
\cmidrule{2-6}          & Aspect & Task & Dataset & \multicolumn{1}{c}{Description} & Metrics \\
\cmidrule{2-6}          & \multirow{19}[8]{*}{\raisebox{-4em}{\rotatebox[origin=c]{90}{Understanding}}} & \multicolumn{1}{c}{\multirow{15}[6]{*}{\makecell[c]{Concept\\Recognition}}} & BC5-disease~\cite{li2016biocreative} & Entity extraction for disease. & \multirow{8}[2]{*}{F1-score} \\
          &       &       & BC5Chem~\cite{li2016biocreative} & Entity extraction for chemical. &  \\
          &       &       & BC4Chem~\cite{savery2020chemical} & Entity extraction for chemical names from PubMed article abstracts. &  \\
          &       &       & Species800~\cite{pafilis2013species} & Extraction of organism names from PubMed article abstracts. &  \\
          &       &       & HoC~\cite{baker2016automatic} & Classification of the hallmarks of cancer given biomedical article abstracts. &  \\
\cmidrule{4-6}          &       &       & HumanDiseaseOntology~\cite{schriml2019human} & Disease ontology-based entity extraction. & \multicolumn{1}{c}{\multirow{2}[2]{*}{\makecell[c]{BLEU, ROUGE,\\AlignScore, Mauve}}} \\
          &       &       & BioLORD~\cite{remy-etal-2023-biolord} & Elaboration of biomedical concepts. &  \\
\cmidrule{4-6}          &       &       & PMC-Patient~\cite{zhao2023large} & Patient-related entity (gender and age for example) extraction from PubMed Central articles. & \multirow{7}[2]{*}{Accuracy} \\
          &       &       & PICO-Participant~\cite{nye2018corpus} & \multirow{3}[0]{*}{\makecell[l]{Information extraction of outcome, interven-\\tion, and participant from article abstracts.}} &  \\
          &       &       & PICO-Intervention~\cite{nye2018corpus} & \multicolumn{1}{c}{} &  \\
          &       &       & PICO-Outcome~\cite{nye2018corpus} & \multicolumn{1}{c}{} &  \\
          &       &       & ADE Corpus~\cite{gurulingappa2012development} & Drug dose extraction given the drug information. &  \\
\cmidrule{3-6}          &       & \multicolumn{1}{c}{\multirow{6}[2]{*}{\makecell[c]{Text\\Summary}}} & MIMIC-IV-Ultrasound~\cite{johnson2023mimic} & \multirow{2}[2]{*}{\makecell[l]{Summarization of patient reports from emer-\\gency departments.}} & \multicolumn{1}{c}{\multirow{6}[2]{*}{\makecell[c]{BLEU, ROUGE,\\AlignScore, Mauve}}} \\
 &       &       & MIMIC-IV-CT~\cite{wallace2021generating} &  &  \\
          &       &       & RCT-Text~\cite{wallace2021generating} & Summarization of medical evidence from clinical studies in literature reviews. &  \\
          &       &       & MedQSum~\cite{lee2021mnlp} & Summarization of patient notes, reports, and health records. &  \\
\cmidrule{2-6}          & \multirow{25}[14]{*}{\raisebox{-4em}{\rotatebox[origin=c]{90}{Reasoning}}} & \multicolumn{1}{c}{\multirow{8}[2]{*}{\makecell[c]{Knowledge\\QA}}} & PubMedQA~\cite{jin2019pubmedqa} & QA data built on PubMed abstracts. & \multirow{8}[2]{*}{Accuracy} \\
          &       &       & MedQA~\cite{jin2021disease} & QA data for medical knowledge assessment. &  \\
          &       &       & MedMCQA~\cite{pal2022medmcqa} & QA data from AIIMS \& NEET PG entrance exams. &  \\
          &       &  & LancetQA~\tablefootnote{\url{https://www.thelancet.com/}} & QA data crawled from Lancet picture quiz gallery. &  \\
          &       &       & NEJMQA~\tablefootnote{\url{https://www.nejm.org/}} & QA and diagnostic challenge requests from NEJM quiz. &  \\
          &       &       & Medbullets~\cite{chen2024benchmarking} & QA data from Medbullets online medical study platform. &  \\
\cmidrule{3-6}          &       & \multicolumn{1}{c}{\multirow{16}[6]{*}{\makecell[c]{Clinical\\Decision Support}}} & DDXPlus~\cite{fansi2022ddxplus} & Diagnostic decision making of synthesized patient data. & \multirow{9}[2]{*}{Accuracy} \\
          &       &       & SEER~\cite{dubey2023using} & Treatment planning for breast cancer cases. &  \\
          &       &       & MIMIC4ED-Hospitalization~\cite{xie2022benchmarking} & \multirow{3}[0]{*}{\makecell[l]{Prediction of clinical outcomes in emergency \\ medicine from MIMIC-IV-ED.}} &  \\
          &       &       & MIMIC4ED-72h ED Revisit~\cite{xie2022benchmarking} & \multicolumn{1}{c}{} &  \\
          &       &       & MIMIC4ED-Critical Triage~\cite{xie2022benchmarking} & \multicolumn{1}{c}{} &  \\
          &       &       & MedNLI-Dis.~\cite{romanov2018lessons} & Discriminative  entailment task for clinical hypotheses. &  \\
          &       &       & PUBHEALTH Ver.~\cite{kotonya2020explainable} & Verification of health-related information from the public. &  \\
\cmidrule{4-6}          &       &       & EBMS~\cite{molla2011development} & Justification verification using the EBMS corpus. & \multicolumn{1}{c}{\multirow{6}[2]{*}{\makecell[c]{BLEU, ROUGE,\\AlignScore, Mauve}}} \\
          &       &       & PUBHEALTH Exp.~\cite{kotonya2020explainable} & Explanation of health-related information from the public. &  \\
          &       &       & ChatDoctor~\cite{li2023chatdoctor} & Patient-doctor dialogues from online medical consultations. &  \\
          &       &       & MedNLI-Gen.~\cite{romanov2018lessons} & Generative entailment task for clinical hypotheses. &  \\
\cmidrule{3-6}
          &       & \multirow{4}[2]{*}{Agent} & AI Hospital~\cite{fan2024ai} & Multi-agent task simulating dynamic medical interactions in Chinese. & \multirow{4}[2]{*}{Accuracy} \\
          &       &       & AgentClinic~\cite{schmidgall2024agentclinic} & Agent benchmark in simulated clinical environments from MedQA and NEJMQA scenarios. &  \\
\cmidrule{3-6}          &       & \multicolumn{1}{c}{\multirow{1}[2]{*}{\makecell[c]{Medical \\ Calculation}}} & MedCalc-Bench~\cite{khandekar2024medcalc} & Medicine dose level calculation from ADE corpus. & Accuracy \\
\cmidrule{2-6}          & \multirow{2}[4]{*}{\raisebox{-1.5em}{\rotatebox[origin=c]{90}{\makecell[c]{Multi-\\linguality}}}} & \multicolumn{1}{c}{\multirow{1}[2]{*}{\makecell[c]{Knowledge\\QA}}} & XMedBench~\cite{wang2024apollo} & Multilingual benchmark for medical understanding and interaction. & Accuracy \\
\cmidrule{3-6}          &       & Agent & AI Hospital~\cite{fan2024ai} & Multi-agent task simulating dynamic medical interactions in Chinese. & Accuracy \\
\cmidrule{2-6}    \end{tabular}%
    }
\end{table}%

First, we present the taxonomy of our evaluation, along with an overview of the evaluation pipeline as shown in Figure~\ref{fig:pipeline}.
Firstly, we specify three aspects of the model capabilities, namely \emph{understanding}, \emph{reasoning}, and \emph{multilinguality}, that correspond to the real-world needs of clinical physicians.
To ensure a comprehensive evaluation, we collect a diverse range of medical tasks and datasets that fall under these three aspects.
Moreover, we explore three prompting strategies in our pipeline, including (1) direct prompting, which instructs LLMs to solve specific problems directly, (2) chain-of-thought, which requires models to think step-by-step before generating the final answer,  (3) few-shot prompting, which providing models with several examples to learn the input-output mapping on the fly.
Lastly, appropriate metrics are utilized to measure the discrepancy between generated responses and ground-truth answers. Details about metrics utilized in each dataset are provided in Table~\ref{table:dataset_scenarios}.

\subsection{Aspects and Tasks}
\label{sec:aspects}
In Table~\ref{table:dataset_scenarios}, our evaluation efforts are structured into three main parts: aspect, task, and dataset. 
Specifically, a \textbf{dataset} refers to the data itself along with the metrics used in the current context. We utilize 35 existing datasets and create 2 additional challenging datasets for evaluation.
A \textbf{task} is a collection of multiple datasets that share a common goal or evaluate similar capabilities within the model. We categorize all 37 datasets into 6 tasks for clearer evaluation and analysis.
An \textbf{aspect} describes a specific capability or property to understand how well the model performs in a particular area.
In our evaluation pipeline, we focus on three key aspects.

Formally, we illustrate these three evaluation aspects with their corresponding tasks as follows:
\begin{itemize} [leftmargin=12pt]
    \item \textbf{Understanding} refers to the model's ability to utilize its internal medical knowledge to comprehend medical concepts. For example, in concept recognition task, the model is required to extract or elaborate medical concepts from article~\cite{savery2020chemical,pafilis2013species,nye2018corpus} or diagnosis report~\cite{zhao2023large}. And in text summarization, the model need to understand concepts in complex texts to generate a concise summary~\cite{lee2021mnlp,wallace2021generating,johnson2019mimic,johnson2023mimic}.
    \item \textbf{Reasoning} is the ability to conduct multiple steps of logical thinking to arrive at the conclusion. In question answering tasks, the model is prompted to select correct option from multi-choices based on reasoning derived from the medical information provided in the question. In addition to common question-answering datasets~\cite{jin2019pubmedqa,pal2022medmcqa,jin2021disease}, we collect real-world clinical questions from The Lancet, the New England Journal of Medicine (NEJM), and Medbullets~\cite{chen2024benchmarking} to better assess the clinical utility of LLMs. In the clinical suggestion task, the model is required to provide treatment suggestions~\cite{dubey2023using,li2023chatdoctor} or diagnostic decisions~\cite{xie2022benchmarking,fansi2022ddxplus} based on patients' information. In the AI Hospital~\cite{fan2024ai} and AgentClinic~\cite{schmidgall2024agentclinic} datasets, we task the model with serving as a medical agent. Furthermore, in the MedCalc-Bench~\cite{khandekar2024medcalc} dataset, the model is required to perform mathematical reasoning and calculate answers.
    \item \textbf{Multilinguality} is the ability to complete a task when the languages of input instruction and/or output answers are changed to different languages. For example, XMedBench~\cite{wang2024apollo} dataset requires LLMs to answer medical questions in six languages, including Chinese, Arabic, Hindi, Spanish, Chinese and English. In AI Hospital dataset~\cite{fan2024ai}, the model is required to serve as an agent using Chinese.
\end{itemize}

\subsection{Metrics}
\label{sec:metrics}
In this section, we elaborate on metrics employed in our evaluation pipeline.
\begin{itemize}[leftmargin=12pt]
    \item \textbf{Accuracy} is used to directly measure the percentage of models' generated answer which exactly match with the ground-truth. 
    We use accuracy for multi-choice question datasets, MedCalc-Bench~\cite{khandekar2024medcalc} dataset, and portions of clinical suggestion and concept recognition datasets where the ground-truth answer is a single word or phrase.
    \item \textbf{F1-score}~\cite{scikit-learn} is the harmonic mean of precision and recall. It is employed in datasets where the model is required to select multiple correct answers.
    \item \textbf{BLEU}~\cite{papineni2002bleu} and \textbf{ROUGE}~\cite{lin2002manual} are NLP metrics measuring the similarity between the generated respond and the ground-truth. Specifically, we utilize BLEU-1 and ROUGE-1 for all free-form generation tasks in our evaluation.
    \item  \textbf{AlignScore}~\cite{zha2023alignscore} is a metric to measure the factual consistency of generated text. In this paper, we use AlignScore for all free-form generation tasks to evaluate the extent of model's hallucination.
    \item \textbf{Mauve}~\cite{pillutla2021mauve} is a measure of gap between distribution of generated and human-written text. It is employed for all free-form generation tasks. 
\end{itemize}
All metrics range from 0 to 100, and a higher number indicates better quality output from the model.

\section{Experiments}
\subsection{Experiment Details}

\paragraph{Prompting strategies.}
For most datasets, we employ the same prompting strategy as described in previous literature~\cite{wu2024towards,nori2023can, nori2023capabilities}: For knowledge QA tasks, agent tasks, medical calculation tasks, and multilingual-related tasks, we use the direct prompting evaluation method, which is consistent with the settings of these benchmarks.
For other tasks derived from MedS-Bench~\cite{wu2024towards}, we follow their benchmark settings, leveraging a few-shot (3-shot) prompt strategy with its template shown in Section~\ref{prompt_format_cot}.
As officially suggested by OpenAI, common prompting techniques such as Chain-of-Thought (CoT)~\cite{wei2022chain} and in-context examples may not boost \texttt{o1}'s performance as it has implicit CoT built in. 
To further validate this claim, we also investigate the effect of several advanced promptings in our evaluation (\eg, CoT, Self-Consistency~\cite{wang2022self}, and Reflex~\cite{shinn2024reflexion}), the detailed input instruction formats are in Section~\ref{prompt_format_cot}

\paragraph{Models for evaluation.}
We choose the following models to evaluate: \texttt{GPT-3.5} (gpt-3.5-turbo-0125)\footnote{\url{https://platform.openai.com/docs/models/gpt-3-5-turbo/}}, an advanced language model by OpenAI known for its enhanced contextual understanding; \texttt{GPT-4} (gpt-4-0125-preview)~\cite{achiam2023gpt}, the successor to \texttt{GPT-3.5} with significant improvements in reasoning and language comprehension; \texttt{o1} (o1-preview-2024-09-12)~\cite{openai-o1}, the lastest LLM model that is capable of performing highly complex reasoning by employing chain-of-thought reasoning.
Apart from these close-source models, we have also incorporated two open-source ones in our experiments: MEDITRON-70B~\cite{chen2023meditron}, an LLM trained with medical-centric data and Llama3-8B~\cite{meta2024introducing}, the latest and strongest open LLM right now.

\begin{table}[t]
  \centering
  \setlength{\tabcolsep}{4pt}
  \caption{\textbf{Accuracy} (Acc.) or \textbf{F1} results on 4 tasks across 2 aspects. Model performances with * are taken from~\citet{wu2024towards} as the reference. We use the gray background to highlight \texttt{o1} results. And we present the average score (Average) of each metric in the table} %
    \resizebox*{!}{.7\textwidth}{
    \begin{tabular}{ccccccccc}
    \toprule
    \multirow{2}{*}{\textbf{Aspect}}     & \multirow{2}{*}{\textbf{Task}}     & \multirow{2}{*}{\textbf{Datasets}} & \multirow{2}{*}{\textbf{Metric}} & {\bl{\texttt{o1}}} & {\makecell{\texttt{GPT-4}}} & {\makecell{\texttt{GPT-3.5}}} & {\makecell[c]{\texttt{MEDITRON*}\\(70B)}} & {\makecell[c]{\texttt{Llama3*}\\(8B)}} \\
    \midrule
    \multirow{5}[0]{*}{\raisebox{-5em}{\rotatebox[origin=c]{90}{\textbf{\footnotesize{~~Understanding}}}}}
    & \multirow{10}[2]{*}{\makecell[c]{Concept\\Recognition}} 
        & PMC-Patient~\cite{zhao2023large} & Acc. & \bl{76.4} & 75.7 & 74.4 &  72.2      & \textbf{96.0} \\
        & & PICO-Participant~\cite{nye2018corpus} & Acc. & \bl{\textbf{75.0}} & \textbf{75.0} & 52.5 &  72.1     & 58.2 \\
        & & PICO-Intervention~\cite{nye2018corpus} & Acc. & \bl{77.5} & 75.0 & 75.0 &  46.6     &  \textbf{79.1}\\
        & & PICO-Outcome~\cite{nye2018corpus} & Acc. & \bl{\textbf{67.5}} & 65.0 & 60.0 &   51.2 & 58.2 \\
        & & ADE Corpus~\cite{gurulingappa2012development} & Acc. & \bl{78.3} & 78.3 & 71.6 &   \textbf{95.7} & 69.6 \\
        & & Average & Acc. & \bl{\textbf{74.9}} & 73.8 & 66.7 & 67.6 & 72.2 \\

        \cmidrule{3-9}
        
        & & BC5-disease~\cite{li2016biocreative} & F1 & \bl{\textbf{69.5}} & 63.0 & 38.9 &    1.4 & 25.3 \\
        & & BC5Chem~\cite{li2016biocreative} & F1 & \bl{\textbf{72.2}} & 71.2 & 43.1 & 4.2 & 37.9 \\
        & & BC4Chem~\cite{savery2020chemical} & F1 & \bl{\textbf{73.4}} & 65.1 & 32.7 & 2.0 & 19.5 \\
        & & Species800~\cite{pafilis2013species} & F1 & \bl{\textbf{71.6}} & 66.8 & 55.4 & 0.4 & 11.9\\
        & & HoC~\cite{pafilis2013species} & F1 & \bl{\textbf{76.3}} & 59.0 & 59.8 & 23.7 & 38.3\\
        & & Average & F1 & \bl{\textbf{72.6}} & 65.0 & 46.0 &  6.3 & 26.6 \\
    \midrule
    \multirow{14}[0]{*}{\raisebox{-3em}{\rotatebox[origin=c]{90}{\textbf{\footnotesize{Reasoning}}}}}
    & \multirow{7}[2]{*}{\makecell[c]{Clinical\\Decision Support}} & DDXPlus~\cite{fansi2022ddxplus} & Acc. & \bl{\textbf{64.0}} & 56.0 & 41.0 &  29.6 & 33.8 \\
        & & SEER~\cite{dubey2023using} & Acc.  & \bl{\textbf{80.0}} & 69.6 & 5.0  &   68.3    & 56.1  \\
        & & \makecell[c]{MIMIC4ED\\-Hospitalization}~\cite{xie2022benchmarking} & Acc. & \bl{\textbf{64.0}} & 61.0 & 62.0 & 56.3 & 39.1  \\
        & & \makecell[c]{MIMIC4ED\\-72h ED Revisit}~\cite{xie2022benchmarking} & Acc. & \bl{\textbf{59.7}} & 58.0 & 53.6 & 48.5 & 9.3 \\
        & & \makecell[c]{MIMIC4ED\\-Critical Triage}~\cite{xie2022benchmarking} & Acc. & \bl{61.7} & \textbf{66.7} & 58.7 & 45.7 & 8.8  \\
        & & MedNLI-Dis.~\cite{romanov2018lessons} & Acc. & \bl{\textbf{88.0}} & 84.0 & 57.0 &  60.9      & 63.9 \\
        & & PUBHEALTH Ver.~\cite{kotonya2020explainable} & Acc. & \bl{\textbf{76.4}} & 75.7 & 74.4 &   32.7    & 63.9 \\
        & & Average & Acc. & \bl{\textbf{70.5}} & 67.3 & 50.2 & 48.9 & 39.3 \\
    \cmidrule{2-9}
    &\multirow{6}[2]{*}{\makecell[c]{Knowledge\\QA}} & PubMedQA~\cite{jin2019pubmedqa} & Acc. & \bl{\textbf{75.0}} & 52.8 & 25.4 & 74.4 &  73.0 \\
        & & MedQA~\cite{jin2021disease} & Acc. & \bl{\textbf{75.5}} & 69.7 & 53.8 & 47.9 & 60.9 \\
        & & MedMCQA~\cite{pal2022medmcqa} & Acc. & \bl{\textbf{95.0}} & 79.5 & 58.8 & 59.2 & 50.7 \\
        & & Medbullets~\cite{chen2024benchmarking} & Acc. &\bl{\textbf{90.6}} & 66.9 & 50.7 & - & - \\
         & & LancetQA & Acc. & \bl{\textbf{81.5}} & 76.0 & 61.0 & - &  - \\
        & & NEJMQA & Acc.  & \bl{\textbf{91.2}} & 83.5 & 65.0 & - & - \\
        & & Average & Acc. & \bl{\textbf{84.8}} & 71.4 & 52.5 & 60.5 & 61.5 \\
    \cmidrule{2-9}
    & Medical Calculation  & MedCalc-Bench~\cite{khandekar2024medcalc} & Acc. & \bl{\textbf{34.9}} & 25.5 & 10.8 &    -   & - \\
    \bottomrule
    \end{tabular}%
    }
  \label{tab:acc}%
\end{table}

\definecolor{baselinecolor}{gray}{.9}

\begin{table}[t]
\centering
\small
\setlength{\tabcolsep}{2pt}
\caption{\textbf{BLEU-1 (B-1)} and \textbf{ROUGE-1 (R-1)} results on 3 tasks across 2 aspects. We use the gray background to highlight \texttt{o1} results. We also present the average score (Average) of each metric}
\resizebox*{!}{.36\textwidth}{
\begin{tabular}{ccccccccccccc}
    \toprule
    \multirow{3}{*}{\textbf{Aspect}} &\multirow{3}{*}{\textbf{Task}} & \multirow{3}{*}{\textbf{Datasets}} & \multicolumn{2}{c}{\bl{\texttt{o1}}} & \multicolumn{2}{c}{\texttt{GPT-4}} & \multicolumn{2}{c}{\texttt{GPT-3.5}} & \multicolumn{2}{c}{\makecell[c]{\texttt{MEDITRON*}\\(70B)}} & \multicolumn{2}{c}{\makecell[c]{\texttt{Llama3*}\\(8B)}} \\

    \cmidrule{4-13} 

    &  &   & \bl{B-1 $\uparrow$}      & \bl{R-1 $\uparrow$}    & B-1 $\uparrow$       & R-1 $\uparrow$       & B-1 $\uparrow$        & R-1 $\uparrow$        & B-1 $\uparrow$        & R-1 $\uparrow$        & B-1 $\uparrow$       & R-1 $\uparrow$       \\
    \midrule
    \multirow{6}[0]{*}{\raisebox{-2.5em}{\rotatebox[origin=c]{90}{\textbf{Understanding}}}}
    &\multirow{4}{*}{\makecell{Text\\Summary}}
    & MIMIC-IV-Ultrasound~\cite{johnson2023mimic} & \bl{\textbf{22.2}} & \bl{\textbf{28.8}} & 15.9 & 27.0 & 11.0 & 21.1 & 3.8 & 6.1 & 18.1 & 20.0 \\
    & & MIMIC-IV-CT~\cite{johnson2023mimic} & \bl{19.0} & \bl{26.4} & 15.7 & 22.7 & 18.7 & 25.9 & 16.3 & 23.9 & \textbf{24.5} & \textbf{29.4} \\
    & & RCT-Text~\cite{wallace2021generating} & \bl{19.5} & \bl{23.4} & 19.5 & 23.4 & \textbf{20.6} & \textbf{24.2} & 4.0 & 16.4 & 15.4 & 14.6 \\
    & & MedQSum~\cite{lee2021mnlp} & \bl{\textbf{39.2}} & \bl{\textbf{46.8}} & 36.3 & 43.0 & 26.5 & 39.6 & 15.6 & 23.1 & 22.5 & 25.1 \\
    & & Average & \bl{\textbf{25.0}} & \bl{\textbf{31.4}} & 21.8 & 29.0 & 19.2 & 27.7 & 9.9 & 17.4 & 20.1 & 22.3 \\

    \cmidrule{2-13} 

    & \multirow{2}{*}{\makecell{Concept\\Recognition}}
    & HumanDO~\cite{schriml2019human} & \bl{\textbf{24.9}} & \bl{\textbf{33.1}} & 9.7 & 16.2 & 12.2 & 19.4 & 7.7 & 25.4 & 14.9 & 18.8 \\
    & & BioLORD~\cite{remy-etal-2023-biolord} & \bl{\textbf{23.0}} & \bl{\textbf{31.8}} & 14.7 & 21.8 & 12.8 & 19.1 & 11.8 & 22.7 & 8.9 & 14.6 \\
    & & Average & \bl{\textbf{24.0}} & \bl{\textbf{32.5}} & 12.2 & 19.0 & 12.5 & 19.3 & 9.8 & 24.1 & 11.9 & 16.7 \\
    
    \midrule

    \multirow{4}[0]{*}{\raisebox{-1.5em}{\rotatebox[origin=c]{90}{\textbf{Reasoning}}}}
    & \multirow{4}{*}{\makecell{Clinical\\Decision Support}}
    & EBMS~\cite{molla2011development} & \bl{\textbf{16.2}} & \bl{\textbf{20.4}} & 12.0 & 16.3 & 15.4 & 19.4 & 11.6 & 15.8 & 16.5 & 16.5 \\
    & & PUBHEALTH Exp.~\cite{kotonya2020explainable} & \bl{15.8} & \bl{\textbf{23.6}} & 15.1 & 22.0 & 16.6 & 23.6 & 6.1 & 8.7 & \textbf{16.8} & {20.3} \\
    & & ChatDoctor~\cite{li2023chatdoctor} & \bl{12.2} & \bl{\textbf{27.6}} & \textbf{20.9} & 4.7 & 14.0 & 27.0 & - & - & - & - \\
    & & MedNLI-Gen.~\cite{romanov2018lessons} & \bl{\textbf{17.0}} & \bl{\textbf{26.0}} & 16.9 & 25.8 & 10.0 & 18.3 & 4.4 & 14.1 & 21.3 & 22.8 \\
    & & Average & \bl{15.3} & \bl{\textbf{24.4}} & \textbf{16.2} & 17.2 & 14.0 & 22.1 & 7.4 & 12.9 & 18.2 & 19.9 \\
\bottomrule
\end{tabular}
}
\label{tab:nlp_metrics}
\end{table}

\subsection{Main Result: \emph{Yes!} We are one Step Closer to an AI Doctor \texorpdfstring{\includegraphics[width=.4cm]{ai_doctor.png}}.}
\paragraph{Enhanced ability of \texttt{o1} transfers to its clinical understanding.}
Given the established results from \texttt{o1}, which underscore its remarkable effectiveness in knowledge and reasoning abilities such as mathematical problem-solving and code generation~\cite{openai-o1}, we observe that this superior capability can also be transferred to the specific clinical knowledge understanding. 
Results presented in Table~\ref{tab:acc} demonstrate that \texttt{o1} outperforms other models on the \emph{understanding} aspect in most clinical tasks. 
We also present these statistics in Figure~\ref{fig:radar}, where we observe that \texttt{o1} has a larger cover radius across various medical datasets.
For instance, on 5 concept recognition datasets that use F1 as the metric, \texttt{o1} outperforms both \texttt{GPT-4} and \texttt{GPT-3.5} by an average of 7.6\% and 26.6\%, respectively (i.e., 72.6\% \vs 65.0\% \vs 46.0\%), with a notable 24.5\% average improvement on the widely used BC4Chem dataset. 

Additionally, on the summarization task in Table~\ref{tab:nlp_metrics}, \texttt{o1} achieves a 2.4\% and 3.7\% increase in ROUGE-1 score over \texttt{GPT-4} and \texttt{GPT-3.5} (\ie, 31.4\% \vs 29.0\% \vs 27.7\%), demonstrating its enhanced capacity for real-world clinical understanding. 
This improved performance confirms that advancements in general NLP capabilities for LLMs can effectively translate to enhanced model understanding in the medical domain.

    \begin{table}[htbp]
\centering
\small
\setlength{\tabcolsep}{5pt}
\caption{\textbf{AlignScore} and \textbf{Mauve} results on 3 tasks across 2 aspects}
\resizebox*{!}{.3\textwidth}{
\begin{tabular}{ccccccccc}
    \toprule
    \multirow{2}{*}{\textbf{Aspect}} & \multirow{2}{*}{\textbf{Task}} & \multirow{2}{*}{\textbf{Datasets}} & \multicolumn{3}{c}{\textbf{AlignScore $\uparrow$}} & \multicolumn{3}{c}{\textbf{Mauve $\uparrow$}}\\
    & & & \bl{\texttt{o1}} & \texttt{GPT-4} & \texttt{GPT-3.5} & \texttt{o1} & \texttt{GPT-4} & \texttt{GPT-3.5}\\

    \midrule

    \multirow{6}{*}{\rotatebox[origin=c]{90}{\textbf{Understanding}}}
    & \multirow{4}{*}{\makecell[c]{Text\\Summary}}
    & MIMIC-IV-Ultrasound~\cite{johnson2023mimic} & \bl{27.5} & \textbf{30.9} & 23.6 & 6.1 & \textbf{7.4} & 7.3 \\
    & & MIMIC-IV-CT~\cite{johnson2023mimic} & \bl{\textbf{14.4}} & 13.3 & 13.8 & 0.4 & \textbf{0.5} & \textbf{0.5} \\
    & & RCT-Text~\cite{wallace2021generating} & \bl{4.9} & 4.9 & \textbf{5.7} & \textbf{3.1} & 2.7 & 11.9 \\
    & & MedQSum~\cite{lee2021mnlp} & \bl{34.5} & \textbf{37.1} & 13.6 & 42.1 & \textbf{52.7} & 0.6 \\
    & & Average & \bl{20.3} & \textbf{21.6} & 14.2 & 12.9 & \textbf{15.8} & 5.1 \\

    \cmidrule{2-9}

    & \multirow{2}{*}{\makecell[c]{Concept\\Recognition}}
    & HumanDO~\cite{schriml2019human} & \bl{\textbf{17.5}} & 5.5 & 5.2 & \textbf{8.2} & 0.4 & 0.4 \\
    & & BioLORD~\cite{remy-etal-2023-biolord} & \bl{13.0} & \textbf{19.0} & 17.9 & \textbf{51.6} & 4.2 & 1.1 \\
    & & Average & \bl{\textbf{15.3}} &  12.3 &  11.6 & \textbf{29.9} & 2.3 & 0.8 \\

    \midrule

    \multirow{4}{*}{\rotatebox[origin=c]{90}{\textbf{Reasoning}}}
    & \multirow{4}{*}{\makecell[c]{Clinical\\Decision Support}}
    & EBMS~\cite{molla2011development} & \bl{\textbf{9.0}} & 6.6 & 5.7 & \textbf{19.5} & 1.9 & 2.3 \\
    & & PUBHEALTH Exp.~\cite{kotonya2020explainable} & \bl{14.8} & \textbf{19.0} & 17.9 & \textbf{2.1} & 0.8 & 1.1 \\
    & & ChatDoctor~\cite{li2023chatdoctor}  & \bl{\textbf{26.5}} & 20.4 & 16.6 & \textbf{0.7} & 0.5 & 0.6 \\
    & & MedNLI-Gen.~\cite{romanov2018lessons} & \bl{6.8} & \textbf{9.7} & 2.5 & \textbf{5.3} & 4.5 & 0.9 \\
    & & Average & \bl{\textbf{14.3}} & 13.9 & 10.7 & \textbf{6.9} & 1.9 & 1.2 \\
    \bottomrule
\end{tabular}
}
\label{tab:alignscore_mauve_adjusted}
\end{table}

\paragraph{The \texttt{o1} model demonstrates strong reasoning in clinical diagnosis scenarios.}
On the reasoning aspect, \texttt{o1} takes a significant step forward in demonstrating its advantages in real-world diagnostic situations. 
In our newly constructed challenging QA tasks, NEJMQA and LacentQA, \texttt{o1} showcases an average accuracy improvement of 8.9\% and 27.1\% over the performance of \texttt{GPT-4} (79.6\%) and \texttt{GPT-3.5} (61.5\%) on the respective datasets (Table~\ref{tab:acc}). 
Another noteworthy improvement in \texttt{o1} is its capacity for mathematical reasoning, elevating the baseline of MedCalc-Bench to 34.9\%, which surpasses \texttt{GPT-4} by a significant 9.4\%. 
In more complex reasoning scenarios that involve multi-turn conversations and environmental simulations, \texttt{o1} outperforms both \texttt{GPT-4} and \texttt{GPT-3.5} on the AgentClinic benchmark, achieving accuracy gains of at least 15.5\% and 10\% with scores of 45.5\% and 20.0\% on its MedQA and NEJM subsets, respectively.
These observations serve as compelling evidence of \texttt{o1}'s competence in complex real-world diagnosis and clinical utility scenarios.

\begin{figure*}[t]
  \centering
  \includegraphics[page=1, width=0.8\textwidth]{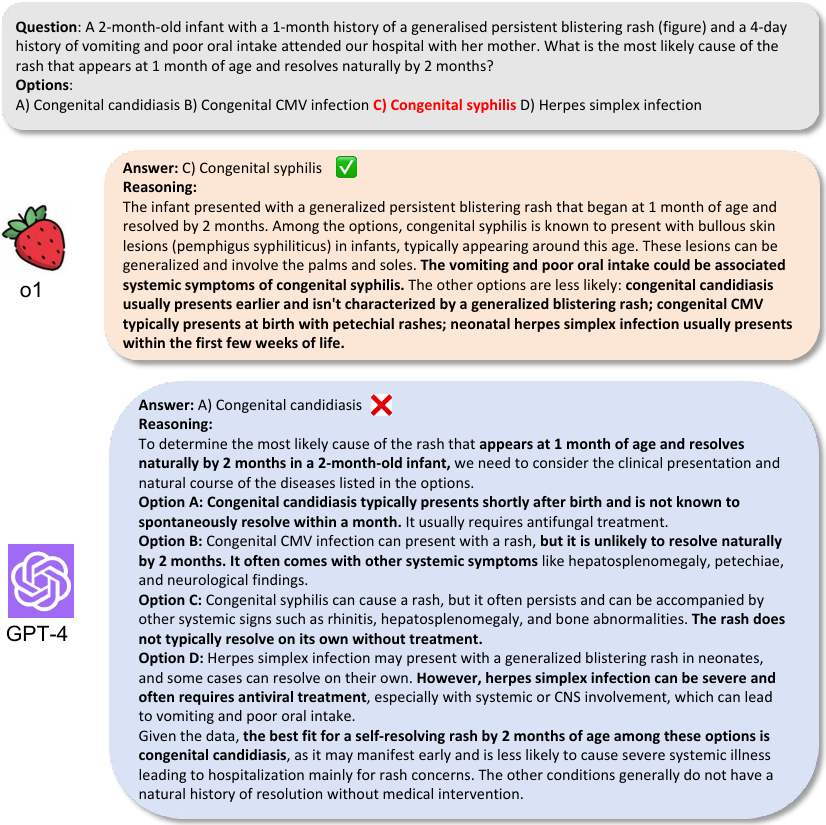}
  \caption{Answers from \texttt{o1} and \texttt{GPT-4} on a question from LancetQA. \texttt{o1} provides a more concise and accurate reasoning process compared to \texttt{GPT-4}.}
  \label{fig:case_2}
\end{figure*}

In addition to delivering higher accuracy, \texttt{o1} provides more concise and straightforward answers. In the example illustrated in Figure~\ref{fig:case_2}, \texttt{o1} generates shorter interpretations while offering the correct answer. 
In contrast, \texttt{GPT-4} tends to generate hallucinated explanations alongside incorrect answers. 
We believe \texttt{o1}'s improvement in both knowledge and reasoning is primarily attributed to the enhanced data and infrastructure employed during the training process (\eg, CoT data and the reinforcement learning technique).

These results together provide a positive answer to the question we raised in this paper: \emph{Yes!} We are getting closer to an automatic AI doctor with the latest \texttt{o1} model.

\begin{table}[t]
\centering
\small
\setlength{\tabcolsep}{5pt}
\caption{\textbf{Accuracy} of LLMs on two agentic benchmarks}
\resizebox*{!}{.18\textwidth}{
\begin{tabular}{cccccc|cc}
\toprule
\multirow{2}{*}{\textbf{Tasks}} & \multicolumn{5}{c}{\makecell[c]{\textbf{AI Hospital}\\\cite{fan2024ai}}}  & \multicolumn{2}{c}{\makecell[c]{\textbf{AgentClinic}\\\cite{schmidgall2024agentclinic}}}  \\
\cmidrule{2-8}
& Symp. & \makecell[c]{Medical\\Exam.} & \makecell[c]{Diagnostic\\Results} & \makecell[c]{Diagnostic\\Rationales} & \makecell[c]{Treatment\\Plan} & MedQA & NEJM    \\
\midrule
\bl{\texttt{o1}}   & \bl{\textbf{67.0}} & \bl{43.4} & \bl{\textbf{45.1}} & \bl{45.1} & \bl{\textbf{39.9}} & \bl{\textbf{45.5}} &  \bl{\textbf{20.0}}  \\
\texttt{GPT-4} & 66.7 & \textbf{45.0} & 44.2 & \textbf{45.8} & 38.2 & 30.4 &  10.0  \\
\texttt{GPT-3.5} & 62.0 & 40.7 & 35.8& 36.3 & 24.7 & 25.2 & 7.5 \\
\bottomrule
\end{tabular}
}
\label{tab:agent_bench}
\end{table}

\subsection{Further Analysis}\label{subsec:further_analysis}
\paragraph{No model excels across all tasks in the medical domain.}
Table~\ref{tab:acc} and Table~\ref{tab:nlp_metrics} indicate that, for now, there are always trade-offs (even under the same metric) to be made when selecting a model to use in the medical domain. 
One example is the clinical decision support task in Table~\ref{tab:acc}, \texttt{o1} outperforms both \texttt{GPT-4} and \texttt{GPT-3.5} on most datasets, but lags far behind \texttt{GPT-4} on the MIMIC4ED-Critical Triage dataset by 5\% in accuracy.
Interestingly, we also found the recent released open LLM---\texttt{Llama3} takes a lead in PMC-Patient and PICO-Intervention datasets with an unexpected 19.6\% accuracy gap between \texttt{o1} and \texttt{Llama3} on PMC-Patient (76.4\% \vs 96.0\%).
Nevertheless, \texttt{o1} comes close to being the best in most situations, it boasts a leading position across datasets in clinical decision support, knowledge QA, and medical calculation. 
This claim is supported by the average result over 19 dataset accuracy in Table~\ref{tab:acc} and Figure~\ref{fig:bar}: 
\texttt{o1} (74.3\%) $>$ \texttt{GPT-4} (68.1\%) $>$ \texttt{GPT-3.5} (53.2\%)

\paragraph{Advanced prompting can partially help models trained with CoT data.}
\texttt{o1} was released using chain-of-thought (CoT) data embedding in the training process; however, we found that applying the CoT prompting still enhances \texttt{o1}'s performance on knowledge QA tasks in medicine, as shown in Table~\ref{tab:cot}. The table reveals an average boost of 3.18\% over the original 83.6\% accuracy of \texttt{o1}. While this improvement is not as significant as with \texttt{GPT-4}, CoT proves to be a promising way for guiding \texttt{o1} in medical tasks.
However, when it comes to other fancy promptings, such as self-consistency (SC)~\cite{wang2022self} and reflex~\cite{shinn2024reflexion}, this conclusion may not stand still.
We witness an average performance decline of 12.8\% using these two strategies compared to only CoT on LancetQA (Table~\ref{tab:prompting_technique}).
\begin{table}[t]
  \centering
  \small
  \setlength{\tabcolsep}{12pt}
  \caption{\textbf{Accuracy} results of model results with/without CoT prompting on 5 knowledge QA datasets}
    \begin{tabular}{ccccc}
    \toprule
    {\textbf{Datasets}} & \texttt{o1}    & \texttt{o1} (CoT) & \texttt{GPT-4} & \texttt{GPT-4} (CoT) \\
    \midrule
          PubMedQA~\cite{jin2019pubmedqa} & 75.0 & \textbf{75.2} & 52.8 & \textbf{62.2} \\
          MedQA~\cite{jin2021disease} & 95.0 & \textbf{95.2} & 79.5 & \textbf{86.1} \\
          MedMCQA~\cite{pal2022medmcqa} & 75.5 & \textbf{81.9} & 69.7 & \textbf{72.6} \\
          LancetQA & 81.5 & \textbf{85.5} & 76.0 & \textbf{81.5} \\
          NEJMQA & 91.2 & \textbf{96.3} & 83.5 & \textbf{86.4} \\
    \bottomrule
    \end{tabular}%
  \label{tab:cot}%
\end{table}

\begin{table}[t]
  \centering
  \setlength{\tabcolsep}{10pt}
  \caption{\textbf{Accuracy} ablation results of using different promptings using \texttt{o1} on our LancetQA}
    \begin{tabular}{ccc|c}
    \toprule
    CoT & SC & Reflex & Accuracy \\
    \midrule
           & & & 81.5 \\
          $\checkmark$ & & & 85.5 \\
          $\checkmark$ & $\checkmark$ &  & 84.5 \\
          $\checkmark$ & & $\checkmark$  & 61.0 \\
    \bottomrule
    \end{tabular}%
  \label{tab:prompting_technique}%
\end{table}

\begin{table}[t!]
\centering
\small
\setlength{\tabcolsep}{5pt}
\caption{\textbf{Accuracy} of models on the multilingual task, XmedBench~\cite{wang2024apollo}}
\begin{tabular}{cccccccc}
\toprule
\textbf{Models} & \textbf{English} & \textbf{Chinese} & \textbf{French} & \textbf{Spanish} & \textbf{Arabic} & \textbf{Hindi} & \textbf{Average} \\
\midrule
\bl{\texttt{o1}} & \bl{\textbf{76.4}} & \bl{\textbf{80.2}} & \bl{\textbf{95.4}} & \bl{\textbf{95.0}} & \bl{\textbf{74.9}} & \bl{\textbf{89.3}} & \bl{\textbf{85.2}} \\
\texttt{GPT-4} & 75.7 & 61.0 & 89.4 & 91.2 & 60.8 & 76.3 & 75.7 \\
\texttt{GPT-3.5} & 72.0 & 47.4 & 58.9 & 74.2 & 39.7 & 32.5 & 54.1 \\
\texttt{Meditron-70B*} & 58.7 & 44.3 & 53.3 & 59.7 & 19.3 & 31.3 & 44.4 \\
\bottomrule
\end{tabular}
\label{tab:multilingual}
\end{table}

\paragraph{Hallucination remains a significant challenge.}
We use AlignScore~\cite{zha2023alignscore} to evaluate hallucination in LLMs. In Table~\ref{tab:alignscore_mauve_adjusted}, the \texttt{o1} model demonstrates a 1.3\% decrease in AlignScore compared to \texttt{GPT-4} across five text summarization datasets. 
Moreover, the overall improvements of \texttt{o1} across three tasks (Table~\ref{tab:alignscore_mauve_adjusted}) in AlignScore significantly lag behind those of other evaluation metrics---averaging 0.7 in AlignScore compared to 9.9 in Mauve relative to \texttt{GPT-4}. This indicates that \texttt{o1} is still susceptible to language hallucination, highlighting that such problem remains a persistent challenge in LLMs.
\paragraph{\texttt{o1} struggles in reasoning over complex multilingual tasks.}
Advanced LLMs are expected to demonstrate equivalent reasoning abilities to languages other than English. 
However, as \texttt{o1} consistently outperforms other models in multilingual QA tasks: \texttt{o1} (85.2\%) $>$ \texttt{GPT-4} (75.7\%) $>$ \texttt{GPT-3.5} (54.1\%) on average (Table~\ref{tab:multilingual}), it falls short in a much more complex Chinese agent benchmark in Table~\ref{tab:agent_bench}---showing a 1.6\% accuracy drop in the medical examinations scenario over \texttt{GPT-4} (43.4\% \vs 45.0\%), leaving its multilingual reasoning in complex situations to be desired. 
This interesting outcome might be attributed to the lack of multilingual CoT data during \texttt{o1}'s training, as learning complex reasoning routes generally requires more efforts than plain instructions in the few-shot paradigm~\cite{kim2023cot,singh2024aya}.
We present a failure example of \texttt{o1} on AI Hospital in Figure~\ref{fig:ai_hospital}. We identified instances of mixed language output in the generation from the doctor, which contribute to the suboptimal performance of \texttt{o1} in this context.

\paragraph{LLMs are facing biased judgement using different metrics.}
Choosing different metrics can lead to varied results of LLM evaluation~\cite{liang2022holistic}, in our experiments, we observe a similar unaligned trend even leveraging traditional NLP metrics such as BLEU-1, ROUGE-1, and Mauve. 
In most cases from Table~\ref{tab:nlp_metrics}, \texttt{o1} surpasses \texttt{GPT-4} in both two traditional reference-based measurements (\ie, BLEU-1, ROUGE-1) on average. 
One exception arises in the BLEU-1 comparison for clinical suggestion tasks. While \texttt{o1} significantly triumph over \texttt{GPT-4} in ROUGE-L (24.4\% \vs 17.2\%), it surprisingly underperforms in BLEU-1: \texttt{o1} (15.3) $<$ \texttt{GPT-4} (16.2). 
When considering Mauve scores, although \texttt{o1} consistently surpasses \texttt{GPT-4} in both averaged BLEU-1 and ROUGE-1 for text summarization tasks, it still falls short by 2.9 points in Mauve, even when evaluated on the same output texts.
A similar anomaly can also be observed in the comparison between accuracy and F1 score. While \texttt{Llama3} significantly outperforms \texttt{o1} in accuracy on two concept recognition datasets, it consistently falls behind \texttt{o1} in F1 on the same cases.
These findings underscore the urgent need to identify or devise more reliable metrics for modern LLMs.

\section{Discussion}\label{sec:discussion}

\paragraph{What adverse impacts does \texttt{o1} bring?}
The model \texttt{o1} has made significant strides in both general NLP and the medical domain---as demonstrated in this paper. But what adverse impacts does \texttt{o1} have on users compared to the previous generations of LLMs?
While embedding the Chain of Thought (CoT) process during generation by default requires more time~\cite{openai-o1}, what exactly distinguishes \texttt{o1} from other OpenAI models? 
In Table~\ref{tab:time}, we see that \texttt{o1} has more than 2$\times$ and 9$\times$ longer decoding time cost on four medical tasks compared to \texttt{GPT-4} and \texttt{GPT-3.5}, respectively (13.18s \vs 6.89s \vs 1.41s). 
This increased decoding time can lead to significant waiting periods when handling complex tasks.

Additionally, \texttt{o1} does not always outperform other models, with inconsistent performance across different tasks. For instance, in the concept recognition task detailed in Table~\ref{tab:acc}, \texttt{o1} underperforms compared to other LLMs on half of the datasets. 
This discrepancy may relate to recent findings suggesting that CoT data is most advantageous in more complex reasoning tasks~\cite{Sprague2024ToCO}. However, in tasks that do not require complex reasoning, such as concept recognition, \texttt{o1} does not have significant advantages over them.

\paragraph{Rethinking evaluation metrics for stronger LLMs.} 
Traditional evaluation metrics like BLEU and ROUGE, which rely on n-gram overlap, have long been criticized for their limitations in capturing the quality of generated text, particularly for LLMs. As a result, using models like GPT-4 as evaluators, \ie, ``LLM-as-a-judge'', has gained popularity for assessing the outputs of other models. However, this approach may not be valid when applied to the most advanced models such as \texttt{o1}, as GPT-4 is even less capable and thus may produce less reliable evaluation. This is especially true for specialized domain like medicine. Therefore, there is a growing need to develop more robust and nuanced evaluation metrics that can better assess the performance of state-of-the-art LLMs in complex scenarios.

\paragraph{Call for reliable prompting techniques for future LLMs.}
As noted in Section~\ref{subsec:further_analysis}, not all advanced prompting techniques positively impact \texttt{o1}'s performance.
As future LLMs like \texttt{o1} may continue to evolve with internal prompts for efficient user instruction, new prompting methods should consider their adaptability to existing strategies. 
One potential exploration could be the integration of two prompting strategies~\cite{wang2022self,zheng2024critic}.

\paragraph{Limitations.}
While we conduct comprehensive evaluations in the medical domain on understanding, reasoning, and multilingual capabilities, there are many other dimensions to consider such as safety~\cite{han2024towards} and we leave them for future work.  
Additionally, we leave more advanced prompting techniques such as retrieval augmented generation (RAG)~\citep{lewis2020retrieval} for future work, which may enhance the factuality and mitigate hallucination. 
It is worth noting that current GPT-like models may still underperform BERT-based specialists in classification tasks~\cite{nori2023can}. However, we focus on GPT-like generalists in this paper due to their greater flexibility as zero-shot learners. 

\section{Conclusion}
This preliminary study assesses 3 important aspects across 35 existing and 2 novel medical datasets using the latest \texttt{o1} model. It marks the first step towards a holistic evaluation of \texttt{o1} in medicine, and we present our initial results, analysis, and discussion over the benchmark. The findings provide convincing evidence that \texttt{o1} is narrowing the gap between AI and human doctors, shaping the vision of an ideal AI doctor closer to reality.

\section*{Acknowledgement}
This work is partially supported by the OpenAI Researcher Access Program and Microsoft Accelerate Foundation Models Research Program. Q.J. is supported by the NIH Intramural Research Program, National Library of Medicine. The content is solely the responsibility of the authors and does not necessarily represent the official views of the funding agencies.

\bibliographystyle{iclr2025_conference}
\bibliography{o1_eval}

\begin{thebibliography}{66}
\providecommand{\natexlab}[1]{#1}
\providecommand{\url}[1]{\texttt{#1}}
\expandafter\ifx\csname urlstyle\endcsname\relax
  \providecommand{\doi}[1]{doi: #1}\else
  \providecommand{\doi}{doi: \begingroup \urlstyle{rm}\Url}\fi

\bibitem[Achiam et~al.(2023)Achiam, Adler, Agarwal, Ahmad, Akkaya, Aleman, Almeida, Altenschmidt, Altman, Anadkat, et~al.]{achiam2023gpt}
Josh Achiam, Steven Adler, Sandhini Agarwal, Lama Ahmad, Ilge Akkaya, Florencia~Leoni Aleman, Diogo Almeida, Janko Altenschmidt, Sam Altman, Shyamal Anadkat, et~al.
\newblock Gpt-4 technical report.
\newblock \emph{arXiv preprint arXiv:2303.08774}, 2023.

\bibitem[Bai et~al.(2023)Bai, Bai, Chu, Cui, Dang, Deng, Fan, Ge, Han, Huang, et~al.]{bai2023qwen}
Jinze Bai, Shuai Bai, Yunfei Chu, Zeyu Cui, Kai Dang, Xiaodong Deng, Yang Fan, Wenbin Ge, Yu~Han, Fei Huang, et~al.
\newblock Qwen technical report.
\newblock \emph{arXiv preprint arXiv:2309.16609}, 2023.

\bibitem[Baker et~al.(2016)Baker, Silins, Guo, Ali, H{\"o}gberg, Stenius, and Korhonen]{baker2016automatic}
Simon Baker, Ilona Silins, Yufan Guo, Imran Ali, Johan H{\"o}gberg, Ulla Stenius, and Anna Korhonen.
\newblock Automatic semantic classification of scientific literature according to the hallmarks of cancer.
\newblock \emph{Bioinformatics}, 32\penalty0 (3):\penalty0 432--440, 2016.

\bibitem[Bubeck et~al.(2023)Bubeck, Chandrasekaran, Eldan, Gehrke, Horvitz, Kamar, Lee, Lee, Li, Lundberg, et~al.]{bubeck2023sparks}
S{\'e}bastien Bubeck, Varun Chandrasekaran, Ronen Eldan, Johannes Gehrke, Eric Horvitz, Ece Kamar, Peter Lee, Yin~Tat Lee, Yuanzhi Li, Scott Lundberg, et~al.
\newblock Sparks of artificial general intelligence: Early experiments with gpt-4.
\newblock \emph{arXiv preprint arXiv:2303.12712}, 2023.

\bibitem[Chen et~al.(2024)Chen, Fang, Singla, and Dredze]{chen2024benchmarking}
Hanjie Chen, Zhouxiang Fang, Yash Singla, and Mark Dredze.
\newblock Benchmarking large language models on answering and explaining challenging medical questions.
\newblock \emph{arXiv preprint arXiv:2402.18060}, 2024.

\bibitem[Chen et~al.(2023)Chen, Cano, Romanou, Bonnet, Matoba, Salvi, Pagliardini, Fan, K{\"o}pf, Mohtashami, et~al.]{chen2023meditron}
Zeming Chen, Alejandro~Hern{\'a}ndez Cano, Angelika Romanou, Antoine Bonnet, Kyle Matoba, Francesco Salvi, Matteo Pagliardini, Simin Fan, Andreas K{\"o}pf, Amirkeivan Mohtashami, et~al.
\newblock Meditron-70b: Scaling medical pretraining for large language models.
\newblock \emph{arXiv preprint arXiv:2311.16079}, 2023.

\bibitem[Dong et~al.(2022)Dong, Li, Dai, Zheng, Wu, Chang, Sun, Xu, and Sui]{dong2022survey}
Qingxiu Dong, Lei Li, Damai Dai, Ce~Zheng, Zhiyong Wu, Baobao Chang, Xu~Sun, Jingjing Xu, and Zhifang Sui.
\newblock A survey on in-context learning.
\newblock \emph{arXiv preprint arXiv:2301.00234}, 2022.

\bibitem[Dubey et~al.(2023)Dubey, Tiwari, Singh, Goldberg, and Pinsky]{dubey2023using}
Snigdha Dubey, Gaurav Tiwari, Sneha Singh, Saveli Goldberg, and Eugene Pinsky.
\newblock Using machine learning for healthcare treatment planning.
\newblock \emph{Frontiers in Artificial Intelligence}, 6:\penalty0 1124182, 2023.

\bibitem[Fan et~al.(2024)Fan, Tang, Chen, Wang, Wei, Xi, Huang, and Zhou]{fan2024ai}
Zhihao Fan, Jialong Tang, Wei Chen, Siyuan Wang, Zhongyu Wei, Jun Xi, Fei Huang, and Jingren Zhou.
\newblock Ai hospital: Interactive evaluation and collaboration of llms as intern doctors for clinical diagnosis.
\newblock \emph{arXiv preprint arXiv:2402.09742}, 2024.

\bibitem[Fansi~Tchango et~al.(2022)Fansi~Tchango, Goel, Wen, Martel, and Ghosn]{fansi2022ddxplus}
Arsene Fansi~Tchango, Rishab Goel, Zhi Wen, Julien Martel, and Joumana Ghosn.
\newblock Ddxplus: A new dataset for automatic medical diagnosis.
\newblock \emph{Advances in neural information processing systems}, 2022.

\bibitem[Feng et~al.(2024)Feng, Zhang, Gu, Ye, He, and Wang]{feng2024towards}
Guhao Feng, Bohang Zhang, Yuntian Gu, Haotian Ye, Di~He, and Liwei Wang.
\newblock Towards revealing the mystery behind chain of thought: a theoretical perspective.
\newblock \emph{Advances in Neural Information Processing Systems}, 36, 2024.

\bibitem[Gurulingappa et~al.(2012)Gurulingappa, Rajput, Roberts, Fluck, Hofmann-Apitius, and Toldo]{gurulingappa2012development}
Harsha Gurulingappa, Abdul~Mateen Rajput, Angus Roberts, Juliane Fluck, Martin Hofmann-Apitius, and Luca Toldo.
\newblock Development of a benchmark corpus to support the automatic extraction of drug-related adverse effects from medical case reports.
\newblock \emph{Journal of biomedical informatics}, 45\penalty0 (5):\penalty0 885--892, 2012.

\bibitem[Han et~al.(2024)Han, Kumar, Agarwal, and Lakkaraju]{han2024towards}
Tessa Han, Aounon Kumar, Chirag Agarwal, and Himabindu Lakkaraju.
\newblock Towards safe large language models for medicine.
\newblock In \emph{ICML 2024 Workshop on Models of Human Feedback for AI Alignment}, 2024.

\bibitem[Jiang et~al.(2023)Jiang, Sablayrolles, Mensch, Bamford, Chaplot, Casas, Bressand, Lengyel, Lample, Saulnier, et~al.]{jiang2023mistral}
Albert~Q Jiang, Alexandre Sablayrolles, Arthur Mensch, Chris Bamford, Devendra~Singh Chaplot, Diego de~las Casas, Florian Bressand, Gianna Lengyel, Guillaume Lample, Lucile Saulnier, et~al.
\newblock Mistral 7b.
\newblock \emph{arXiv preprint arXiv:2310.06825}, 2023.

\bibitem[Jin et~al.(2021)Jin, Pan, Oufattole, Weng, Fang, and Szolovits]{jin2021disease}
Di~Jin, Eileen Pan, Nassim Oufattole, Wei-Hung Weng, Hanyi Fang, and Peter Szolovits.
\newblock What disease does this patient have? a large-scale open domain question answering dataset from medical exams.
\newblock \emph{Applied Sciences}, 2021.

\bibitem[Jin et~al.(2019)Jin, Dhingra, Liu, Cohen, and Lu]{jin2019pubmedqa}
Qiao Jin, Bhuwan Dhingra, Zhengping Liu, William~W Cohen, and Xinghua Lu.
\newblock Pubmedqa: A dataset for biomedical research question answering.
\newblock \emph{arXiv preprint arXiv:1909.06146}, 2019.

\bibitem[Johnson et~al.(2019)Johnson, Lungren, Peng, Lu, Mark, Berkowitz, and Horng]{johnson2019mimic}
Alistair Johnson, Matt Lungren, Yifan Peng, Zhiyong Lu, Roger Mark, Seth Berkowitz, and Steven Horng.
\newblock Mimic-cxr-jpg-chest radiographs with structured labels.
\newblock \emph{PhysioNet}, 2019.

\bibitem[Johnson et~al.(2023)Johnson, Bulgarelli, Shen, Gayles, Shammout, Horng, Pollard, Hao, Moody, Gow, et~al.]{johnson2023mimic}
Alistair~EW Johnson, Lucas Bulgarelli, Lu~Shen, Alvin Gayles, Ayad Shammout, Steven Horng, Tom~J Pollard, Sicheng Hao, Benjamin Moody, Brian Gow, et~al.
\newblock Mimic-iv, a freely accessible electronic health record dataset.
\newblock \emph{Scientific data}, 10\penalty0 (1):\penalty0 1, 2023.

\bibitem[Khandekar et~al.(2024)Khandekar, Jin, Xiong, Dunn, Applebaum, Anwar, Sarfo-Gyamfi, Safranek, Anwar, Zhang, et~al.]{khandekar2024medcalc}
Nikhil Khandekar, Qiao Jin, Guangzhi Xiong, Soren Dunn, Serina~S Applebaum, Zain Anwar, Maame Sarfo-Gyamfi, Conrad~W Safranek, Abid~A Anwar, Andrew Zhang, et~al.
\newblock Medcalc-bench: Evaluating large language models for medical calculations.
\newblock \emph{arXiv preprint arXiv:2406.12036}, 2024.

\bibitem[Kim et~al.(2023)Kim, Joo, Kim, Jang, Ye, Shin, and Seo]{kim2023cot}
Seungone Kim, Se~June Joo, Doyoung Kim, Joel Jang, Seonghyeon Ye, Jamin Shin, and Minjoon Seo.
\newblock The cot collection: Improving zero-shot and few-shot learning of language models via chain-of-thought fine-tuning.
\newblock \emph{arXiv preprint arXiv:2305.14045}, 2023.

\bibitem[Kotonya \& Toni(2020)Kotonya and Toni]{kotonya2020explainable}
Neema Kotonya and Francesca Toni.
\newblock Explainable automated fact-checking for public health claims.
\newblock \emph{arXiv preprint arXiv:2010.09926}, 2020.

\bibitem[Lee et~al.(2021)Lee, Dang, Uzuner, and Henry]{lee2021mnlp}
Jooyeon Lee, Huong Dang, Ozlem Uzuner, and Sam Henry.
\newblock Mnlp at mediqa 2021: fine-tuning pegasus for consumer health question summarization.
\newblock In \emph{Proceedings of the 20th Workshop on Biomedical Language Processing}, pp.\  320--327, 2021.

\bibitem[Lewis et~al.(2020)Lewis, Perez, Piktus, Petroni, Karpukhin, Goyal, K{\"u}ttler, Lewis, Yih, Rockt{\"a}schel, et~al.]{lewis2020retrieval}
Patrick Lewis, Ethan Perez, Aleksandra Piktus, Fabio Petroni, Vladimir Karpukhin, Naman Goyal, Heinrich K{\"u}ttler, Mike Lewis, Wen-tau Yih, Tim Rockt{\"a}schel, et~al.
\newblock Retrieval-augmented generation for knowledge-intensive nlp tasks.
\newblock \emph{Advances in Neural Information Processing Systems}, 33:\penalty0 9459--9474, 2020.

\bibitem[Li et~al.(2016)Li, Sun, Johnson, Sciaky, Wei, Leaman, Davis, Mattingly, Wiegers, and Lu]{li2016biocreative}
Jiao Li, Yueping Sun, Robin~J Johnson, Daniela Sciaky, Chih-Hsuan Wei, Robert Leaman, Allan~Peter Davis, Carolyn~J Mattingly, Thomas~C Wiegers, and Zhiyong Lu.
\newblock Biocreative v cdr task corpus: a resource for chemical disease relation extraction.
\newblock \emph{Database}, 2016, 2016.

\bibitem[Li et~al.(2023)Li, Li, Zhang, Dan, Jiang, and Zhang]{li2023chatdoctor}
Yunxiang Li, Zihan Li, Kai Zhang, Ruilong Dan, Steve Jiang, and You Zhang.
\newblock Chatdoctor: A medical chat model fine-tuned on a large language model meta-ai (llama) using medical domain knowledge.
\newblock \emph{Cureus}, 15\penalty0 (6), 2023.

\bibitem[Liang et~al.(2022)Liang, Bommasani, Lee, Tsipras, Soylu, Yasunaga, Zhang, Narayanan, Wu, Kumar, et~al.]{liang2022holistic}
Percy Liang, Rishi Bommasani, Tony Lee, Dimitris Tsipras, Dilara Soylu, Michihiro Yasunaga, Yian Zhang, Deepak Narayanan, Yuhuai Wu, Ananya Kumar, et~al.
\newblock Holistic evaluation of language models.
\newblock \emph{arXiv preprint arXiv:2211.09110}, 2022.

\bibitem[Li{\'e}vin et~al.(2024)Li{\'e}vin, Hother, Motzfeldt, and Winther]{lievin2024can}
Valentin Li{\'e}vin, Christoffer~Egeberg Hother, Andreas~Geert Motzfeldt, and Ole Winther.
\newblock Can large language models reason about medical questions?
\newblock \emph{Patterns}, 2024.

\bibitem[Lin \& Hovy(2002)Lin and Hovy]{lin2002manual}
Chin-Yew Lin and Eduard Hovy.
\newblock Manual and automatic evaluation of summaries.
\newblock In \emph{Proceedings of the ACL-02 workshop on automatic summarization}, pp.\  45--51, 2002.

\bibitem[McCarthy et~al.(1955)McCarthy, Minsky, Rochester, and Shannon]{mccarthy2006proposal}
John McCarthy, Marvin~L Minsky, Nathaniel Rochester, and Claude~E Shannon.
\newblock A proposal for the dartmouth summer research project on artificial intelligence, august 31, 1955.
\newblock \emph{AI magazine}, 1955.

\bibitem[Meta(2024)]{meta2024introducing}
AI~Meta.
\newblock Introducing meta llama 3: The most capable openly available llm to date.
\newblock \emph{Meta AI}, 2024.

\bibitem[Moll{\'a} \& Santiago-Martinez(2011)Moll{\'a} and Santiago-Martinez]{molla2011development}
Diego Moll{\'a} and Maria~Elena Santiago-Martinez.
\newblock Development of a corpus for evidence based medicine summarisation.
\newblock In \emph{Proceedings of the Australasian Language Technology Association Workshop 2011}, pp.\  86--94. Australian Language Technology Association, 2011.

\bibitem[Nori et~al.(2023{\natexlab{a}})Nori, King, McKinney, Carignan, and Horvitz]{nori2023capabilities}
Harsha Nori, Nicholas King, Scott~Mayer McKinney, Dean Carignan, and Eric Horvitz.
\newblock Capabilities of gpt-4 on medical challenge problems.
\newblock \emph{arXiv preprint arXiv:2303.13375}, 2023{\natexlab{a}}.

\bibitem[Nori et~al.(2023{\natexlab{b}})Nori, Lee, Zhang, Carignan, Edgar, Fusi, King, Larson, Li, Liu, et~al.]{nori2023can}
Harsha Nori, Yin~Tat Lee, Sheng Zhang, Dean Carignan, Richard Edgar, Nicolo Fusi, Nicholas King, Jonathan Larson, Yuanzhi Li, Weishung Liu, et~al.
\newblock Can generalist foundation models outcompete special-purpose tuning? case study in medicine.
\newblock \emph{arXiv preprint arXiv:2311.16452}, 2023{\natexlab{b}}.

\bibitem[Nye et~al.(2018)Nye, Li, Patel, Yang, Marshall, Nenkova, and Wallace]{nye2018corpus}
Benjamin Nye, Junyi~Jessy Li, Roma Patel, Yinfei Yang, Iain~J Marshall, Ani Nenkova, and Byron~C Wallace.
\newblock A corpus with multi-level annotations of patients, interventions and outcomes to support language processing for medical literature.
\newblock In \emph{Proceedings of the conference. Association for Computational Linguistics. Meeting}, volume 2018, pp.\  197. NIH Public Access, 2018.

\bibitem[OpenAI(2024)]{openai-o1}
OpenAI.
\newblock Openai o1 system card.
\newblock \url{https://openai.com/index/openai-o1-system-card/}, September 2024.

\bibitem[Ouyang et~al.(2022)Ouyang, Wu, Jiang, Almeida, Wainwright, Mishkin, Zhang, Agarwal, Slama, Ray, et~al.]{ouyang2022training}
Long Ouyang, Jeffrey Wu, Xu~Jiang, Diogo Almeida, Carroll Wainwright, Pamela Mishkin, Chong Zhang, Sandhini Agarwal, Katarina Slama, Alex Ray, et~al.
\newblock Training language models to follow instructions with human feedback.
\newblock \emph{Advances in neural information processing systems}, 2022.

\bibitem[Pafilis et~al.(2013)Pafilis, Frankild, Fanini, Faulwetter, Pavloudi, Vasileiadou, Arvanitidis, and Jensen]{pafilis2013species}
Evangelos Pafilis, Sune~P Frankild, Lucia Fanini, Sarah Faulwetter, Christina Pavloudi, Aikaterini Vasileiadou, Christos Arvanitidis, and Lars~Juhl Jensen.
\newblock The species and organisms resources for fast and accurate identification of taxonomic names in text.
\newblock \emph{PloS one}, 8\penalty0 (6):\penalty0 e65390, 2013.

\bibitem[Pal et~al.(2022)Pal, Umapathi, and Sankarasubbu]{pal2022medmcqa}
Ankit Pal, Logesh~Kumar Umapathi, and Malaikannan Sankarasubbu.
\newblock Medmcqa: A large-scale multi-subject multi-choice dataset for medical domain question answering.
\newblock In \emph{Conference on health, inference, and learning}. PMLR, 2022.

\bibitem[Papineni et~al.(2002)Papineni, Roukos, Ward, and Zhu]{papineni2002bleu}
Kishore Papineni, Salim Roukos, Todd Ward, and Wei-Jing Zhu.
\newblock Bleu: a method for automatic evaluation of machine translation.
\newblock In \emph{Proceedings of the 40th annual meeting of the Association for Computational Linguistics}, pp.\  311--318, 2002.

\bibitem[Pedregosa et~al.(2011)Pedregosa, Varoquaux, Gramfort, Michel, Thirion, Grisel, Blondel, Prettenhofer, Weiss, Dubourg, Vanderplas, Passos, Cournapeau, Brucher, Perrot, and Duchesnay]{scikit-learn}
F.~Pedregosa, G.~Varoquaux, A.~Gramfort, V.~Michel, B.~Thirion, O.~Grisel, M.~Blondel, P.~Prettenhofer, R.~Weiss, V.~Dubourg, J.~Vanderplas, A.~Passos, D.~Cournapeau, M.~Brucher, M.~Perrot, and E.~Duchesnay.
\newblock Scikit-learn: Machine learning in {P}ython.
\newblock \emph{Journal of Machine Learning Research}, 12:\penalty0 2825--2830, 2011.

\bibitem[Peng et~al.(2024)Peng, Goldstein, Anthony, Albalak, Alcaide, Biderman, Cheah, Du, Ferdinan, Hou, Kazienko, GV, Kocoń, Koptyra, Krishna, au2, Muennighoff, Obeid, Saito, Song, Tu, Woźniak, Zhang, Zhao, Zhao, Zhou, Zhu, and Zhu]{rwkv6}
Bo~Peng, Daniel Goldstein, Quentin Anthony, Alon Albalak, Eric Alcaide, Stella Biderman, Eugene Cheah, Xingjian Du, Teddy Ferdinan, Haowen Hou, Przemysław Kazienko, Kranthi~Kiran GV, Jan Kocoń, Bartłomiej Koptyra, Satyapriya Krishna, Ronald McClelland~Jr. au2, Niklas Muennighoff, Fares Obeid, Atsushi Saito, Guangyu Song, Haoqin Tu, Stanisław Woźniak, Ruichong Zhang, Bingchen Zhao, Qihang Zhao, Peng Zhou, Jian Zhu, and Rui-Jie Zhu.
\newblock Eagle and finch: Rwkv with matrix-valued states and dynamic recurrence.
\newblock In \emph{COLM}, 2024.

\bibitem[Pillutla et~al.(2021)Pillutla, Swayamdipta, Zellers, Thickstun, Welleck, Choi, and Harchaoui]{pillutla2021mauve}
Krishna Pillutla, Swabha Swayamdipta, Rowan Zellers, John Thickstun, Sean Welleck, Yejin Choi, and Zaid Harchaoui.
\newblock Mauve: Measuring the gap between neural text and human text using divergence frontiers.
\newblock \emph{Advances in Neural Information Processing Systems}, 34:\penalty0 4816--4828, 2021.

\bibitem[Remy et~al.(2024)Remy, Demuynck, and Demeester]{remy-etal-2023-biolord}
François Remy, Kris Demuynck, and Thomas Demeester.
\newblock {BioLORD-2023: semantic textual representations fusing large language models and clinical knowledge graph insights}.
\newblock \emph{Journal of the American Medical Informatics Association}, 2024.

\bibitem[Romanov \& Shivade(2018)Romanov and Shivade]{romanov2018lessons}
Alexey Romanov and Chaitanya Shivade.
\newblock Lessons from natural language inference in the clinical domain.
\newblock \emph{arXiv preprint arXiv:1808.06752}, 2018.

\bibitem[Saunders et~al.(2022)Saunders, Yeh, Wu, Bills, Ouyang, Ward, and Leike]{saunders2022self}
William Saunders, Catherine Yeh, Jeff Wu, Steven Bills, Long Ouyang, Jonathan Ward, and Jan Leike.
\newblock Self-critiquing models for assisting human evaluators.
\newblock \emph{arXiv preprint arXiv:2206.05802}, 2022.

\bibitem[Savery et~al.(2020)Savery, Rogers, Pillai, Mork, and Demner-Fushman]{savery2020chemical}
Max~E Savery, Willie~J Rogers, Malvika Pillai, James~G Mork, and Dina Demner-Fushman.
\newblock Chemical entity recognition for medline indexing.
\newblock \emph{AMIA Summits on Translational Science Proceedings}, 2020:\penalty0 561, 2020.

\bibitem[Schmidgall et~al.(2024)Schmidgall, Ziaei, Harris, Reis, Jopling, and Moor]{schmidgall2024agentclinic}
Samuel Schmidgall, Rojin Ziaei, Carl Harris, Eduardo Reis, Jeffrey Jopling, and Michael Moor.
\newblock Agentclinic: a multimodal agent benchmark to evaluate ai in simulated clinical environments.
\newblock \emph{arXiv preprint arXiv:2405.07960}, 2024.

\bibitem[Schriml et~al.(2019)Schriml, Mitraka, Munro, Tauber, Schor, Nickle, Felix, Jeng, Bearer, Lichenstein, et~al.]{schriml2019human}
Lynn~M Schriml, Elvira Mitraka, James Munro, Becky Tauber, Mike Schor, Lance Nickle, Victor Felix, Linda Jeng, Cynthia Bearer, Richard Lichenstein, et~al.
\newblock Human disease ontology 2018 update: classification, content and workflow expansion.
\newblock \emph{Nucleic acids research}, 47\penalty0 (D1):\penalty0 D955--D962, 2019.

\bibitem[Shinn et~al.(2024)Shinn, Cassano, Gopinath, Narasimhan, and Yao]{shinn2024reflexion}
Noah Shinn, Federico Cassano, Ashwin Gopinath, Karthik Narasimhan, and Shunyu Yao.
\newblock Reflexion: Language agents with verbal reinforcement learning.
\newblock \emph{Advances in Neural Information Processing Systems}, 2024.

\bibitem[Singh et~al.(2024)Singh, Vargus, Dsouza, Karlsson, Mahendiran, Ko, Shandilya, Patel, Mataciunas, OMahony, et~al.]{singh2024aya}
Shivalika Singh, Freddie Vargus, Daniel Dsouza, B{\"o}rje~F Karlsson, Abinaya Mahendiran, Wei-Yin Ko, Herumb Shandilya, Jay Patel, Deividas Mataciunas, Laura OMahony, et~al.
\newblock Aya dataset: An open-access collection for multilingual instruction tuning.
\newblock \emph{arXiv preprint arXiv:2402.06619}, 2024.

\bibitem[Sprague et~al.(2024)Sprague, Yin, Rodriguez, Jiang, Wadhwa, Singhal, Zhao, Ye, Mahowald, and Durrett]{Sprague2024ToCO}
Zayne Sprague, Fangcong Yin, Juan~Diego Rodriguez, Dongwei Jiang, Manya Wadhwa, Prasann Singhal, Xinyu Zhao, Xi~Ye, Kyle Mahowald, and Greg Durrett.
\newblock To cot or not to cot? chain-of-thought helps mainly on math and symbolic reasoning.
\newblock \emph{arXiv preprint arXiv:2409.12183}, 2024.

\bibitem[Touvron et~al.(2023{\natexlab{a}})Touvron, Lavril, Izacard, Martinet, Lachaux, Lacroix, Rozi{\`e}re, Goyal, Hambro, Azhar, et~al.]{touvron2023llama}
Hugo Touvron, Thibaut Lavril, Gautier Izacard, Xavier Martinet, Marie-Anne Lachaux, Timoth{\'e}e Lacroix, Baptiste Rozi{\`e}re, Naman Goyal, Eric Hambro, Faisal Azhar, et~al.
\newblock Llama: Open and efficient foundation language models.
\newblock \emph{arXiv preprint arXiv:2302.13971}, 2023{\natexlab{a}}.

\bibitem[Touvron et~al.(2023{\natexlab{b}})Touvron, Martin, Stone, Albert, Almahairi, Babaei, Bashlykov, Batra, Bhargava, Bhosale, et~al.]{touvron2023llama2}
Hugo Touvron, Louis Martin, Kevin Stone, Peter Albert, Amjad Almahairi, Yasmine Babaei, Nikolay Bashlykov, Soumya Batra, Prajjwal Bhargava, Shruti Bhosale, et~al.
\newblock Llama 2: Open foundation and fine-tuned chat models.
\newblock \emph{arXiv preprint arXiv:2307.09288}, 2023{\natexlab{b}}.

\bibitem[Tu et~al.(2023{\natexlab{a}})Tu, Cui, Wang, Zhou, Zhao, Han, Zhou, Yao, and Xie]{tu2023how}
Haoqin Tu, Chenhang Cui, Zijun Wang, Yiyang Zhou, Bingchen Zhao, Junlin Han, Wangchunshu Zhou, Huaxiu Yao, and Cihang Xie.
\newblock How many unicorns are in this image? a safety evaluation benchmark for vision llms.
\newblock \emph{arXiv preprint arXiv:2311.16101}, 2023{\natexlab{a}}.

\bibitem[Tu et~al.(2023{\natexlab{b}})Tu, Zhao, Wei, and Xie]{tu2023sight}
Haoqin Tu, Bingchen Zhao, Chen Wei, and Cihang Xie.
\newblock Sight beyond text: Multi-modal training enhances llms in truthfulness and ethics.
\newblock \emph{arXiv preprint arXiv:2309.07120}, 2023{\natexlab{b}}.

\bibitem[Wallace et~al.(2021)Wallace, Saha, Soboczenski, and Marshall]{wallace2021generating}
Byron~C Wallace, Sayantan Saha, Frank Soboczenski, and Iain~J Marshall.
\newblock Generating (factual?) narrative summaries of rcts: Experiments with neural multi-document summarization.
\newblock \emph{AMIA Summits on Translational Science Proceedings}, 2021:\penalty0 605, 2021.

\bibitem[Wang et~al.(2023)Wang, Liu, Xi, Qiang, Zhao, Qin, and Liu]{wang2023huatuo}
Haochun Wang, Chi Liu, Nuwa Xi, Zewen Qiang, Sendong Zhao, Bing Qin, and Ting Liu.
\newblock Huatuo: Tuning llama model with chinese medical knowledge.
\newblock \emph{arXiv preprint arXiv:2304.06975}, 2023.

\bibitem[Wang et~al.(2024)Wang, Chen, Chen, Hu, Wang, Wu, Gao, Wan, Li, and Wang]{wang2024apollo}
Xidong Wang, Nuo Chen, Junyin Chen, Yan Hu, Yidong Wang, Xiangbo Wu, Anningzhe Gao, Xiang Wan, Haizhou Li, and Benyou Wang.
\newblock Apollo: Lightweight multilingual medical llms towards democratizing medical ai to 6b people.
\newblock \emph{arXiv preprint arXiv:2403.03640}, 2024.

\bibitem[Wang et~al.(2022)Wang, Wei, Schuurmans, Le, Chi, Narang, Chowdhery, and Zhou]{wang2022self}
Xuezhi Wang, Jason Wei, Dale Schuurmans, Quoc Le, Ed~Chi, Sharan Narang, Aakanksha Chowdhery, and Denny Zhou.
\newblock Self-consistency improves chain of thought reasoning in language models.
\newblock \emph{arXiv preprint arXiv:2203.11171}, 2022.

\bibitem[Wei et~al.(2022)Wei, Wang, Schuurmans, Bosma, Xia, Chi, Le, Zhou, et~al.]{wei2022chain}
Jason Wei, Xuezhi Wang, Dale Schuurmans, Maarten Bosma, Fei Xia, Ed~Chi, Quoc~V Le, Denny Zhou, et~al.
\newblock Chain-of-thought prompting elicits reasoning in large language models.
\newblock \emph{Advances in neural information processing systems}, 2022.

\bibitem[Wu et~al.(2024{\natexlab{a}})Wu, Lin, Zhang, Zhang, Xie, and Wang]{wu2024pmc}
Chaoyi Wu, Weixiong Lin, Xiaoman Zhang, Ya~Zhang, Weidi Xie, and Yanfeng Wang.
\newblock Pmc-llama: toward building open-source language models for medicine.
\newblock \emph{Journal of the American Medical Informatics Association}, pp.\  ocae045, 2024{\natexlab{a}}.

\bibitem[Wu et~al.(2024{\natexlab{b}})Wu, Qiu, Liu, Gu, Li, Zhang, Wang, and Xie]{wu2024towards}
Chaoyi Wu, Pengcheng Qiu, Jinxin Liu, Hongfei Gu, Na~Li, Ya~Zhang, Yanfeng Wang, and Weidi Xie.
\newblock Towards evaluating and building versatile large language models for medicine.
\newblock \emph{arXiv preprint arXiv:2408.12547}, 2024{\natexlab{b}}.

\bibitem[Xie et~al.(2022)Xie, Zhou, Lee, Tan, Li, Rajnthern, Chee, Chakraborty, Wong, Dagan, et~al.]{xie2022benchmarking}
Feng Xie, Jun Zhou, Jin~Wee Lee, Mingrui Tan, Siqi Li, Logasan~S/O Rajnthern, Marcel~Lucas Chee, Bibhas Chakraborty, An-Kwok~Ian Wong, Alon Dagan, et~al.
\newblock Benchmarking emergency department prediction models with machine learning and public electronic health records.
\newblock \emph{Scientific Data}, 9\penalty0 (1):\penalty0 658, 2022.

\bibitem[Zha et~al.(2023)Zha, Yang, Li, and Hu]{zha2023alignscore}
Yuheng Zha, Yichi Yang, Ruichen Li, and Zhiting Hu.
\newblock Alignscore: Evaluating factual consistency with a unified alignment function.
\newblock In \emph{The 61st Annual Meeting Of The Association For Computational Linguistics}, 2023.

\bibitem[Zhao et~al.(2023)Zhao, Jin, Chen, Peng, and Yu]{zhao2023large}
Zhengyun Zhao, Qiao Jin, Fangyuan Chen, Tuorui Peng, and Sheng Yu.
\newblock A large-scale dataset of patient summaries for retrieval-based clinical decision support systems.
\newblock \emph{Scientific data}, 10\penalty0 (1):\penalty0 909, 2023.

\bibitem[Zheng et~al.(2024)Zheng, Lou, Cao, Wen, Ji, Lin, Lu, Han, Zhang, and Sun]{zheng2024critic}
Xin Zheng, Jie Lou, Boxi Cao, Xueru Wen, Yuqiu Ji, Hongyu Lin, Yaojie Lu, Xianpei Han, Debing Zhang, and Le~Sun.
\newblock Critic-cot: Boosting the reasoning abilities of large language model via chain-of-thoughts critic.
\newblock \emph{arXiv preprint arXiv:2408.16326}, 2024.

\end{thebibliography}
\newpage
\appendix

\section{Supplemental Material}
\subsection{Prompting Strategies}

\begin{tcolorbox}[
    title={Base Prompt for MCQ.},
    colback=blue!5!white,
    colframe=blue!40!white,
    left=1mm, right=1mm, top=1mm, bottom=1mm,
    width=\textwidth,
    center,
    fonttitle=\small,
    fontupper=\small,
    label=prompt_base
]

{
    Question: \\
    \{question\} \\
    
    Options: \\
~~~~~~~~A) ...... \\
~~~~~~~~B) ...... \\
\quad\quad....... \\

    \{Format Constraint\}
}
\end{tcolorbox}

\begin{tcolorbox}[
    title={Format Constraint Examples for MCQ.},
    colback=blue!5!white,
    colframe=blue!40!white,
    left=1mm, right=1mm, top=1mm, bottom=1mm,
    width=\textwidth,
    center,
    fonttitle=\small,
    fontupper=\small,
    label=prompt_format_base
]
{
    \textbf{Default:} \\
    Answer only with the option index such as A/B/C/D in plain text.

    \textbf{True/False Statement Questions:} \\
    Answer only with Yes/No in plain text.
}
\end{tcolorbox}
\begin{tcolorbox}[
    title={Few-Shot Prompt.},
    colback=blue!5!white,
    colframe=blue!40!white,
    left=1mm, right=1mm, top=1mm, bottom=1mm,
    width=\textwidth,
    center,
    fonttitle=\small,
    fontupper=\small,
    label=prompt_format_few_shot
]
{
    Case1: ... \\
    Case2: ... \\
    Case3: ... \\
    ... \\
    \{Manually Written Definitions\} \\
    Please learn from the few-shot cases to see what content you have to output.
        \\
    \{Input Case\}
}
\end{tcolorbox}

\begin{tcolorbox}[
    title={CoT Format Constraint.},
    colback=blue!5!white,
    colframe=blue!40!white,
    left=1mm, right=1mm, top=1mm, bottom=1mm,
    width=\textwidth,
    center,
    fonttitle=\small,
    fontupper=\small,
    label=prompt_format_cot
]
{
    Reason step-by-step before answering. \{Base Format Instruction\}. Your final output should strictly follow this format: \\
    \textlangle Reason\textrangle \{your step-by-step reasoning\}\textlangle /Reason\ \textrangle \textlangle Answer\textrangle\{your answer\}\textlangle /Answer\textrangle
}
\end{tcolorbox}

\begin{tcolorbox}[
    title={Self Consistency.},
    colback=blue!5!white,
    colframe=blue!40!white,
    left=1mm, right=1mm, top=1mm, bottom=1mm,
    width=\textwidth,
    center,
    fonttitle=\small,
    fontupper=\small,
    label=prompt_sc
]
{
    Given the following question and the \{n\_sample\} answers, please select the most consistent response with other answers and the question. \{Base Format Constraint\} in strictly this format: \textlangle Answer\textrangle\{your final answer\}\textlangle/Answer\textrangle.\\

    \# Question: \{Base Prompt with CoT\}\\

    \# Answer 1:\\
    \{Model Answer 1\}\\

    \# Answer 2:\\
    \{Model Answer 2\}\\

    \# Answer 3:\\
    \{Model Answer 3\}\\
}

\end{tcolorbox}

\begin{tcolorbox}[
    title={Prompt for Critic Generation for Reflex.},
    colback=blue!5!white,
    colframe=blue!40!white,
    left=1mm, right=1mm, top=1mm, bottom=1mm,
    width=\textwidth,
    center,
    fonttitle=\small,
    fontupper=\small,
    label=prompt_reflex1
]
{
    \{Base Prompt with CoT Format Constraint\}\\

    \# Response:\\
    \{Model Response\}\\

    Please review the answer above and criticize on where might be wrong. If you are absolutely sure it is correct, output `True'.
}
\end{tcolorbox}

\begin{tcolorbox}[
    title={Prompt for Reflected Answer Generation for Reflex.},
    colback=blue!5!white,
    colframe=blue!40!white,
    left=1mm, right=1mm, top=1mm, bottom=1mm,
    width=\textwidth,
    center,
    fonttitle=\small,
    fontupper=\small,
    label=prompt_reflex2
]
{
    \{Base Prompt with CoT Format Constraint\}\\

    \# Original Answer:\\
    \{Model Answer\}\\

    \# Critic:\\
    \{Model Critic\}\\

    Given previous attempts and feedback, carefully consider where you could go wrong in your latest attempt. Using insights from previous attempts, try to solve the task better.
}
\end{tcolorbox}

\begin{tcolorbox}[
    title={Prompt for Final Answer Generation for Reflex.},
    colback=blue!5!white,
    colframe=blue!40!white,
    left=1mm, right=1mm, top=1mm, bottom=1mm,
    width=\textwidth,
    center,
    fonttitle=\small,
    fontupper=\small,
    label=prompt_reflex3
]
{
    \{Base Prompt with CoT Format Constraint\}\\

    \# Answer 1:\\
    \{Reflected Answer 1\}\\

    \# Answer 2:\\
    \{Reflected Answer 2\}\\

    \# Answer 3:\\
    \{Reflected Answer 3\}\\

    Please summarize the previous attempts and feedback and provide a final answer. \{Base Format Constraint\} in strictly this format: \textlangle Answer\textrangle\{your final answer\}\textlangle/Answer\textrangle.
}
\end{tcolorbox}

\subsection{Details about Datasets}
In this paper, we present a summary of 36 medical-related datasets spanning 6 distinct tasks, as outlined in Table~\ref{table:dataset_scenarios}. Notably, the inclusion of commercial models, particularly \texttt{o1}, leads to significant costs and response latency. To address this, for some tasks we randomly sampled a subset of test cases, which are detailed below.

\paragraph{Concept Recognition}
\begin{itemize}[leftmargin=12pt]
    \item \textbf{BC4Chem}~\cite{savery2020chemical} is a dataset comprising 10,000 PubMed abstracts with 84,355 chemical entity mentions, manually annotated by expert chemistry literature curators. The task is to extract chemical names from the given abstracts. For evaluation, we randomly sample 300 instances from the test set.
    
    \item \textbf{BC5Chem} and \textbf{BC5Disease} are from BC5CDR~\cite{li2016biocreative}, a widely-used resource in biomedical natural language processing, annotated for chemical and disease entities and their relationships. Following MedS-Bench~\cite{wu2024towards}, BC5CDR is split into 2 datasets: chemical name extraction and disease name extraction. For evaluation, we randomly sample 300 instances from each task's test set.
    
    \item \textbf{Species800}~\cite{pafilis2013species} comprises 800 PubMed abstracts with annotated organism mentions. The task is to extract organism names from the given abstracts. For evaluation, we randomly sample 300 instances from the test set.
    
    \item \textbf{HoC}~\cite{baker2016automatic} is a specialized dataset containing 1,852 PubMed publication abstracts, expertly annotated according to a taxonomy of cancer hallmarks. The task is to classify the hallmarks of cancer based on the given biomedical publication abstracts. For evaluation, we use the entire test set consisting of 158 instances.
    
    \item \textbf{HumanDiseaseOntology}~\cite{schriml2019human} is a database providing consistent, reusable, and sustainable descriptions of human disease terms, phenotype characteristics, and related medical vocabularies. The task is to explain specified medical professional entities, with the database descriptions serving as ground truth. For evaluation, we randomly sample 300 instances.
    
    \item \textbf{BioLORD}~\cite{remy-etal-2023-biolord} comprises pairs of biomedical concept names and descriptions. The task is to elaborate on concise concepts by generating long, detailed definitions. For evaluation, we randomly sample 300 instances.
    
    \item \textbf{PMC-Patient}~\cite{zhao2023large} is a collection of 167,000 patient summaries extracted from case reports in PubMed Central (PMC), annotated with basic patient information. The task is to extract patient gender and age information from given clinical texts. For evaluation, we randomly sample 300 instances.
    
    \item \textbf{PICO-Participant}, \textbf{PICO-Intervention} and \textbf{PICO-Outcome} are three datasets derived from PICO~\cite{nye2018corpus}, consisting of 5,000 abstracts from medical articles on randomized controlled clinical trials. The tasks involve extracting information about study participants, interventions, and outcomes from given sentences. For evaluation, we use the entire test set of 43 instances for each task.
    
    \item \textbf{ADE Corpus}~\cite{gurulingappa2012development} provides information on drugs and their corresponding adequate doses within sentences. The task is to extract the dosage levels of specified drugs from given sentences and drug names. We use the dataset prompted by Super-Instruction with a 9:1 ratio for instruction tuning and evaluation. The test set consists of 23 instances.
    
\end{itemize}

\paragraph{Text Summary}
\begin{itemize}[leftmargin=12pt]
    \item \textbf{MIMIC-IV-CT} and \textbf{MIMIC-IV-Ultrasound}~\cite{johnson2023mimic,wallace2021generating} are subsets of MIMIC-IV Report, a large deidentified medical dataset of patients admitted to the Beth Israel Deaconess Medical Center. The task is to summarize radiology reports, treating the impression part as a general summary of the findings. Following~\cite{wu2024towards}, we randomly sampled 500 cases from body region part of Chest CT and   100 cases from ultrasound modality for evaluation.
    
    \item \textbf{RCT-Text}~\cite{wallace2021generating} is a dataset for summarizing medical evidence from clinical studies in literature reviews. The task is to output the primary conclusions of each study given the titles and abstracts. For evaluation, we randomly sample 100 instances.
    
    \item \textbf{MedQSum}~\cite{lee2021mnlp} is derived from a large database of de-identified health-related data. The task is to generate a summary of detailed findings from imaging diagnostic reports, with the conclusion of the note serving as ground truth. For evaluation, we randomly sample 100 instances.
\end{itemize}

\paragraph{Knowledge QA} 
\begin{itemize}[leftmargin=12pt]
    \item \textbf{MedQA}~\cite{jin2021disease} is a collection of medical multiple-choice questions in English. We use the 4-option English version with the official split. The test set contains 1273 samples.
    
    \item \textbf{PubMedQA}~\cite{jin2019pubmedqa} is an English question-answering dataset based on PubMed abstracts. The task is to answer research questions with yes/no/maybe. We use the PQA-L subset as the test set, containing 1000 samples.
    
    \item \textbf{MedMCQA}~\cite{pal2022medmcqa} is a large-scale English multiple-choice question-answering dataset from AIIMS \& NEET PG entrance exams. We use the official test split containing 4183 questions, each with 4 choices.
    
    \item \textbf{LancetQA} and \textbf{NEJMQA} are datasets curated from The Lancet and the New England Journal of Medicine case challenges, focusing on patient diagnosis based on symptoms. We use 200 samples for LancetQA and 100 samples for NEJMQA.
    
    \item \textbf{Medbullets}~\cite{chen2024benchmarking} is a dataset curated from the Medbullets online platform, comprising 308 USMLE Step 2\&3 style questions. Each question includes a case description, four answer choices, and an explanation.
\end{itemize}

\paragraph{Clinical Decision Support}
\begin{itemize}[leftmargin=12pt]
    \item \textbf{DDXPlus}~\cite{fansi2022ddxplus} is a dataset for Automatic Symptom Detection and Automatic Diagnosis systems, featuring synthesized patient data. The task is to make diagnostic decisions based on dialogues. For evaluation, we randomly sample 300 instances.
    
    \item \textbf{SEER}~\cite{dubey2023using} is a treatment planning dataset based on the Surveillance, Epidemiology, and End Results breast cancer databases. The task is to recommend treatment plans from five types. For evaluation, we randomly sample 300 instances.
    
    \item \textbf{MIMIC4ED-Hospitalization}, \textbf{MIMIC4ED-72h ED Revisit}, and \textbf{MIMIC4ED-Critical Triage} are datasets from the MIMIC4ED Benchmark~\cite{xie2022benchmarking} for predicting clinical outcomes in emergency medicine. For each dataset, we randomly sample 300 instances for evaluation.
    
    \item \textbf{MedNLI-Dis.} (Discriminative) and \textbf{MedNLI-Gen.} (Generative) are derived from MedNLI~\cite{romanov2018lessons}, a natural language inference dataset for the clinical domain. The dataset involve discriminative and generative entailment based on clinical premises. For each task, we randomly sample 300 instances for evaluation.
    
    \item \textbf{EBMS}~\cite{molla2011development} is a justification verification dataset. We use the entire test set of 304 instances for evaluation.
    
    \item \textbf{PUBHEALTH Exp.} (Explanation)~\cite{kotonya2020explainable} requires models to provide explanations for specified claims using supporting material from given paragraphs. For evaluation, we randomly sample 300 instances.
    
    \item \textbf{PUBHEALTH Ver.} (Verification)~\cite{kotonya2020explainable} is a fact verification task where models determine if a claim contradicts evidence in a given paragraph. For evaluation, we randomly sample 300 instances.
    
    \item \textbf{Chatdoctor}~\cite{li2023chatdoctor} is based on 100K patient-physician conversations from an online medical consultation website\footnote{www.healthcaremagic.com}. The task involves engaging in medical consultations based on this data. 
\end{itemize}
\paragraph{Agent}
\begin{itemize}[leftmargin=12pt]
   \item \textbf{AI Hospital}~\cite{fan2024ai} is a multi-agent framework simulating medical interactions in Chinese. It includes Patient, Examiner, Chief Physician, and Doctor agents, with 506 cases from diverse departments. The task involves simulating clinical scenarios through dialogue. Evaluation uses Chief Physician's 1-4 scale scoring across five dimensions: symptoms, examinations, diagnostic results, rationales, and treatment plan. 200 cases are sampled for evaluation.

   \item \textbf{AgentClinic}~\cite{schmidgall2024agentclinic} is a clinical environment benchmark with 107 patient agents from MedQA and 15 multimodal agents from NEJM challenges. The task is patient diagnosis through dialogue and data collection. Evaluation considers diagnostic accuracy and patient perception metrics in biased scenarios.
\end{itemize}

\paragraph{Medical Calculation}
\begin{itemize}[leftmargin=12pt]
\item \textbf{MedCalc-Bench}~\cite{khandekar2024medcalc} evaluates LLMs' medical calculation abilities using 1,047 instances across 55 tasks. It requires computing medical values from patient notes and questions. Evaluation compares LLM outputs to ground truth, with exact matches for rule-based and 5\% tolerance for equation-based calculators.
\end{itemize}

\paragraph{Multilinguality}
\begin{itemize}[leftmargin=12pt]
\item \textbf{XMedBench}~\cite{wang2024apollo} is a multilingual medical benchmark in six languages: English, Chinese, Hindi, Spanish, French, and Arabic. It uses multiple-choice questions from various sources, including translated versions for Arabic and Hindi. The task evaluates LLMs' medical knowledge across languages, using accuracy as the primary metric.
\item \textbf{AI Hospital}~\cite{fan2024ai} is a multi-agent framework simulating medical interactions in Chinese. We also include this dataset into the multilinguality aspect because it is in Chinese.
\end{itemize}
\subsection{Model-based Evaluation}
As discussed in Section~\ref{sec:discussion}, Rethinking evaluation metrics for stronger LLMs, we also explore using techniques such as "LLM-as-a-judge" to assess the quality of generated outputs. Table ~\ref{tab:gpt_evaluation} shows that \texttt{o1} achieves nearly the same score as \texttt{GPT-4} and outperforms \texttt{GPT-3.5} (i.e., 3.3\% vs. 3.3\% vs. 3.0\%), which contrasts with the traditional evaluation metrics in Table~\ref{tab:nlp_metrics}. This indicates that the ``LLM-as-a-judge'' method may be unreliable when applied to advanced models like \texttt{o1}, as GPT-4, being less capable, may provide less accurate evaluations. This limitation is particularly evident in specialized domains such as medicine. The prompt used for "LLM-as-a-judge" is shown in Figure~\ref{prompt_judge}.

\begin{table}[htbp]
\centering
\setlength{\tabcolsep}{5pt}
\caption{GPT Evaluation Score Comparison}
\begin{tabular}{c|c|ccc}
\toprule
\multirow{2}{*}{\textbf{Task}} & \multirow{2}{*}{\textbf{Datasets}} & \multicolumn{3}{c}{\textbf{GPT Score} $\uparrow$} \\
& & \bl{\texttt{o1}} & \texttt{GPT-4} & \texttt{GPT-3.5} \\
\midrule
\multirow{4}{*}{\textbf{Text Summarization}}
& medqsum & \bl{4.1} & 3.8 & 4.1 \\
& RCT-Text & \bl{3.2} & 3.2 & 3.1 \\
& MIMIC-IV-Ultrasound & \bl{3.8} & 3.8 & 3.4 \\
& MIMIC-IV-CT & \bl{3.8} & 3.8 & 3.7 \\
\midrule
\multirow{6}{*}{\textbf{Clinical Suggestion}}
& MedNLI-Generative & \bl{2.3} & 2.4 & 2.5 \\
& EMBS Justification Ver. & \bl{3.1} & 3.0 & 3.0 \\
& PUBHEALTH Exp. & \bl{3.0} & 3.3 & 3.2 \\
& Do Entity Exp. & \bl{3.7} & 3.6 & 3.3 \\
& BioLORD Concept Exp. & \bl{3.3} & 3.3 & 3.0 \\
& ChatDoctor & \bl{2.5} & 2.6 & -- \\
\midrule
\multicolumn{2}{c|}{\textbf{Average}} & \bl{3.3} & 3.3 & 3.3 \\
\bottomrule
\end{tabular}
\label{tab:gpt_evaluation}
\end{table}

\begin{tcolorbox}[
    title={Prompt for LLM-as-a-judge.},
    colback=blue!5!white,
    colframe=blue!40!white,
    left=1mm, right=1mm, top=1mm, bottom=1mm,
    width=\textwidth,
    center,
    fonttitle=\small,
    fontupper=\small,
    label=prompt_judge
]
{

    You are a senior medical expert. Please evaluate the quality of the medical text material provided by medical interns based on the expert medical text material as a reference answer. The quality is divided into five levels: \\
    
    5. The assistant result completely matches the reference. \\
    4. The assistant result is generally consistent with the reference, with only a small part of omissions or errors. \\
    3. The assistant result partially matches the reference, but there are some omissions and errors. \\
    2. The assistant result is mostly inconsistent with the reference, with many omissions and errors. \\
    1. The assistant result is completely inconsistent with the reference. \\

    \{Input Medical Questions\} \\
    Assistant Result: \{Result\} \\
    Reference Answer: \{Reference\} \\
\\
Please note: \\
(1) Focus on the factual content of the medical answers, without concern for style, grammar, punctuation, and non-medical content.
(2) Your response should be in the format.
Rating: (int)
}
\end{tcolorbox}
\subsection{Decoding Time}
\begin{table}[htbp]
\centering
\setlength{\tabcolsep}{1pt}
\caption{Model \textbf{time} cost and averaged number of decoding tokens for 4 datasets across 4 tasks}
\resizebox*{!}{.43\textwidth}{
\begin{tabular}{cccccccc}
\toprule
\textbf{Task} & \textbf{Dataset} & \textbf{Model} & \textbf{Time (s)} & \textbf{\makecell[c]{Prompt\\Tokens}} & \textbf{\makecell[c]{Completion\\Tokens}} & \textbf{\makecell[c]{Reasoning\\Tokens}} & \textbf{\makecell[c]{Total\\Tokens}} \\
\midrule
\multirow{3}{*}{\makecell[c]{Knowledge\\QA}} & \multirow{3}{*}{MedQA}
& \texttt{o1} & 11.13 & 247.78 & 953.42 & 924.16 & 1201.20 \\
& & \texttt{GPT-4} & 0.83 & 236.20 & 9.26 & 0 & 245.46 \\
& & \texttt{GPT-3.5} & 0.52 & 236.20 & 10.02 & 0 & 246.22 \\
\midrule
\multirow{3}{*}{\makecell[c]{Clinical\\ Decision Support}} & \multirow{3}{*}{ChatDoctor}
& \texttt{o1} & 11.40 & 122.64 & 1127.44 & 83.84 & 1250.08 \\
& & \texttt{GPT-4} & 18.88 & 124.24 & 509.28 & 0 & 633.52 \\
& & \texttt{GPT-3.5} & 2.40 & 124.24 & 150.10 & 0 & 274.34 \\
\midrule
\multirow{3}{*}{\makecell[c]{Text\\Summary}} & \multirow{3}{*}{MIMIC-IV}
& \texttt{o1} & 20.56 & 1305.54 & 1080.54 & 1057.28 & 1373.32 \\
& & \texttt{GPT-4} & 6.26 & 1254.84 & 162.68 & 0 & 1417.52 \\
& & \texttt{GPT-3.5} & 2.02 & 1254.84 & 159.94 & 0 & 1414.78 \\
\midrule
\multirow{3}{*}{\makecell[c]{Concept\\Recognition}} & \multirow{3}{*}{BC5Chem}
& \texttt{o1} & 9.62 & 292.78 & 1080.54 & 1057.28 & 1373.32 \\
& & \texttt{GPT-4} & 1.60 & 297.24 & 19.64 & 0 & 316.88 \\
& & \texttt{GPT-3.5} & 0.68 & 297.24 & 12.80 & 0 & 310.04 \\
\bottomrule
\end{tabular}
}
\label{tab:time}
\end{table}

We evaluated the model's time cost and the average number of decoding tokens across various tasks, including Knowledge QA, Clinical Decision Support, Text Summary, and Concept Recognition. For each task, we select a representative dataset and perform inference on 50 samples. The time and decoded tokens are then averaged to obtain the results for each response, as illustrated in Table. ~\ref{tab:time}. The decoding time for \texttt{o1} is significantly higher than both \texttt{GPT-4} and \texttt{GPT-3.5}, taking more than double the time of \texttt{GPT-4} and over nine times that of \texttt{GPT-3.5} across four medical tasks (13.18s compared to 6.89s and 1.41s, respectively).

\subsection{Case Study}
\begin{figure*}[t]
  \centering
  \includegraphics[page=1, width=0.9\textwidth]{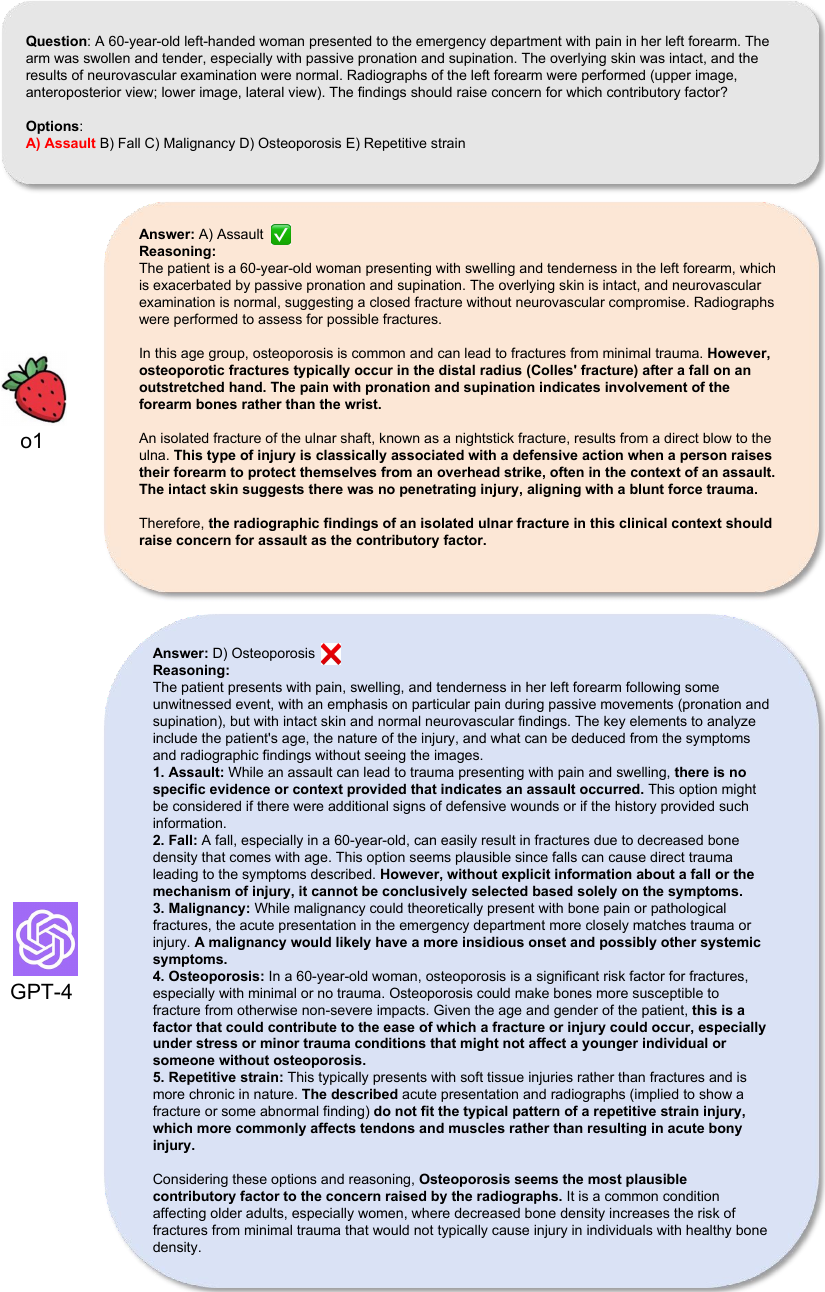}
  \caption{Comparison of the answers from GPT-o1 and GPT-4 for a question from NEJMQA.  o1 provides a more concise and accurate reasoning process compared to GPT-4.}
  \label{fig:case_1}
\end{figure*}

\begin{figure*}[t]
  \centering
  \includegraphics[height=0.95\textheight, width=0.95\textwidth]{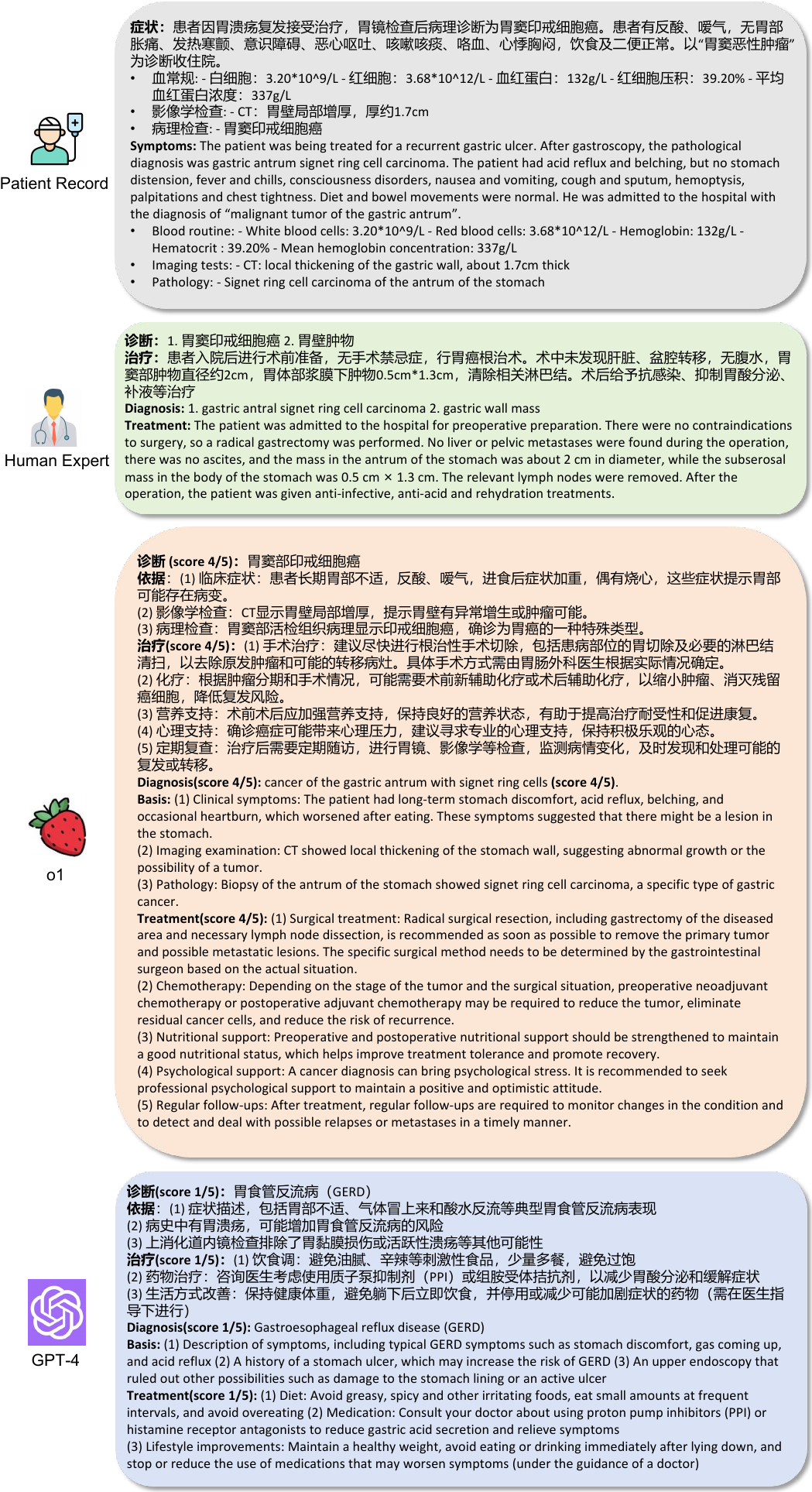}
  \caption{Comparison of the answers from GPT-o1 and GPT-4 for a case from the Chinese dataset AI Hospital, along with its English translation. o1 offers a more precise diagnosis and practical treatment suggestions compared to GPT-4.}
  \label{fig:case_3}
\end{figure*}

To demonstrate how the advanced \texttt{o1} model outperforms previous methods in potential clinical applications, we present comparative case studies. As shown in Figure ~\ref{fig:case_1} and Figure ~\ref{fig:case_2}, when responding to questions from NEJMQA and LancetQA, \texttt{o1} provides the correct answer with a more concise and precise reasoning process. In contrast, \texttt{GPT-4} generates a longer reasoning path while leading to an incorrect answer. in Figure ~\ref{fig:case_3}, when offering diagnosis and treatment based on patient records, \texttt{o1} delivers a more accurate diagnosis and practical treatment plan, closely aligning with human experts and outperforming \texttt{GPT-4}. These cases demonstrate that \texttt{o1} more closely resembles an AI Doctor.

\begin{figure*}[t]
  \centering
  \includegraphics[page=1, width=0.9\textwidth]{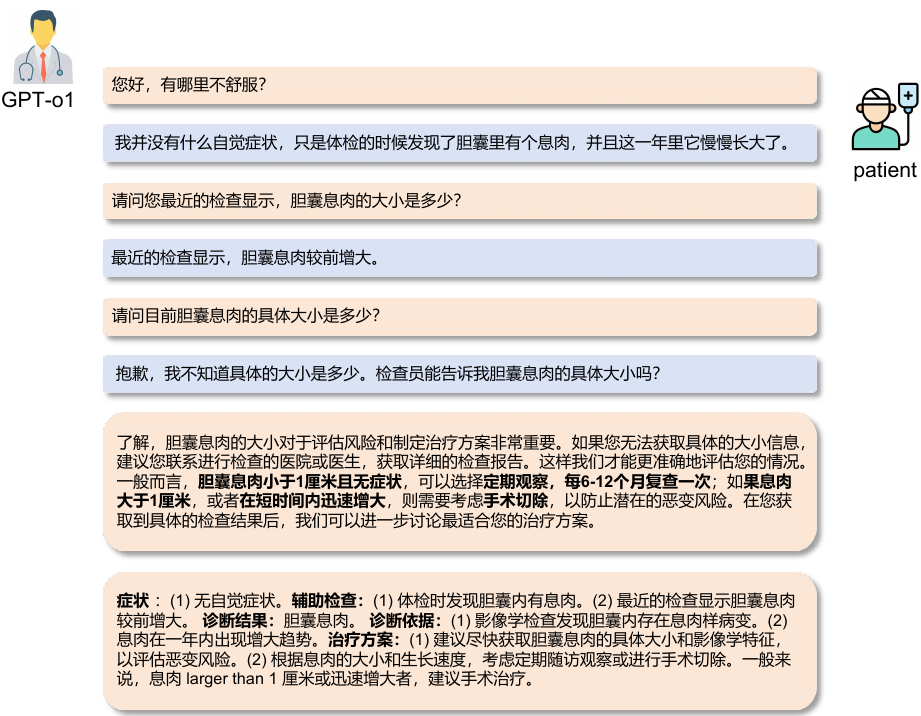}
  \caption{\textbf{Failure case} of \texttt{o1} on AI Hospital. The model struggles with generating the right diagnosis and outputs mixed-language, resulting to its suboptimal performance in this context.}
  \label{fig:ai_hospital}
\end{figure*}

\end{document}